\documentclass{article}
\usepackage{amsmath}  

\usepackage{PRIMEarxiv}

\usepackage[utf8]{inputenc} 
\usepackage[T1]{fontenc}    
\usepackage{hyperref}       
\usepackage{url}            
\usepackage{booktabs}       
\usepackage{amsfonts}       
\usepackage{nicefrac}       
\usepackage{microtype}      
\usepackage{lipsum}
\usepackage{fancyhdr}       
\usepackage{graphicx}       
\graphicspath{{media/}}     
\usepackage{tcolorbox} 
\usepackage{float} 
\usepackage{graphicx}
\usepackage{subcaption}
\usepackage{tabularx}  
\usepackage{booktabs}  
\usepackage{array}     

\title{Comparative Analysis and Parametric Tuning of PPO, GRPO, and DAPO for LLM Reasoning Enhancement
}

\author{
  Yongsheng Lian \\
  Mechanical Engineering Department \\
  University of Louisville \\
  Louisville, KY 40223\\
  \texttt{yongsheng.lian@louisville.edu}
}

\begin{document}
\maketitle

\begin{abstract}

This study presents a systematic comparison of three Reinforcement Learning (RL) algorithms—PPO, GRPO, and DAPO—for improving complex reasoning in large language models (LLMs). Our main contribution is a controlled transfer-learning evaluation: models are first fine-tuned on the specialized Countdown Game and then assessed on a suite of general-purpose reasoning benchmarks. Across all tasks, RL-trained models outperform their corresponding base models, although the degree of improvement differs by benchmark.

Our parametric analysis offers practical guidance for RL-based LLM training. Increasing the group size in GRPO and DAPO leads to more stable training dynamics and higher accuracy, while the impact of the KL-penalty coefficient is non-monotonic. Additionally, we find that the Dynamic Sampling (DS) component in DAPO does not improve performance; in fact, the best overall results are achieved with DAPO when DS is disabled.

\end{abstract}


\section{Introduction}

Reinforcement learning  has demonstrated remarkable effectiveness in enhancing large language model reasoning. Notable examples include OpenAI's O1 model~\cite{openai2024reasoning} and DeepSeek’s R1 model~\cite{guo2025deepseek}. Both show significant improvements in chain-of-thought reasoning after RL training. These models can break down a complex problem into simpler steps, recognize and correct mistakes, and adopt alternative strategies when initial attempts fail. These success shows that RL can mmake LLMs think more productively and systematically, and boost LLM's problem-solving capabilities.

A variety of RL-based methods have emerged in recent literature, each tailored to address specific challenges in LLM fine-tuning. Proximal Policy Optimization (PPO) remains a foundational method~\cite{schulman2017proximalpolicyoptimizationalgorithms}, widely used for its balance of stability and efficiency. Group Relative Policy Optimization (GRPO)~\cite{shao2024deepseekmathpushinglimitsmathematical} refines PPO by estimating advantages relative to peer generations, enhancing reasoning capabilities without requiring a value function. The Decoupled Clip and Dynamic sAmpling Policy Optimization (DAPO)~\cite{yu2025dapo} builds upon GRPO by introducing several key improvements.  It achieved state-of-the-art performance on the AIME 2024 benchmark. 

Each of these methods has been conducted on different models and datasets, making direct comparisons challenging. It remains unclear how their improvements scale across the same underlying architecture or data. This motivates our work: by applying PPO, GRPO, and DAPO to the same model and training dataset, we can systematically evaluate their relative performance, isolate the contributions of each RL strategy, and provide clearer guidance for future LLM fine-tuning efforts. 

The remainder of this paper is structured as follows: Section 2 formalizes the objectives and core differences of PPO, GRPO, and DAPO. Section 3 details a comprehensive parametric study. Section 4 presents the comparative performance results across standard benchmarks, and Section 5 concludes with a summary of key findings.

\begin{figure}
  \centering
  \includegraphics[width=0.8\textwidth]{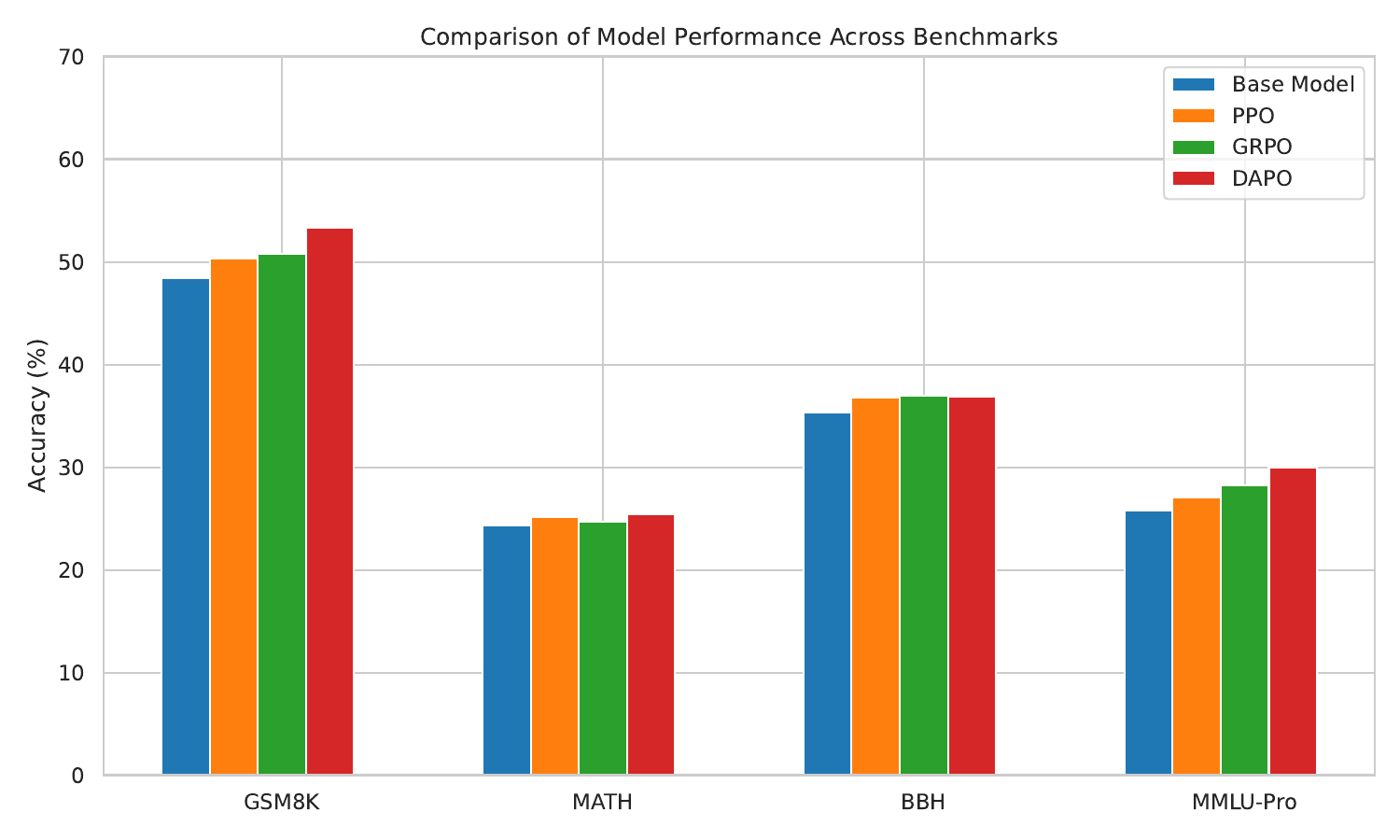}  
  \caption{Model performance on different benchmarks. DAPO with no dynamic sampling.}
  \label{fig:completion_length}
\end{figure}

\section{Method}

While PPO~\cite{schulman2017proximalpolicyoptimizationalgorithms}. serves as the de facto standard for RLHF, its dependence on a learned value function (critic) introduces memory overhead and training instability. GRPO~\cite{shao2024deepseekmathpushinglimitsmathematical} and its variant DAPO~\cite{yu2025dapo} attempt to mitigate this by eliminating the critic in favor of group-relative advantage estimation. In this section, we formalize the objectives of these three algorithms to highlight their structural differences in loss aggregation and regularization.

\subsection{Proximal Policy Optimization}

PPO aims to enhance a policy as effectively as possible using current data, while avoiding updates that could inadvertently degrade performance. Unlike TRPO, which depends on a complex second-order optimization, PPO leverages straightforward first-order methods along with mechanisms that gently constrain policy changes~\cite{schulman2017proximalpolicyoptimizationalgorithms}. This approach simplifies implementation and delivers performance that matches TRPO.

TRPO maximizes a surrogate objective with a constraint. PPO achieves similar performance benefits as TRPO while avoiding the complexities of a constrained optimization by using a simple clipped surrogate objective:
 
\begin{equation}\label{eq:ppo_objective}
\mathcal{J}^{\text{CLIP}}(\theta) = \mathbb{E}_t \Big[ \min \Big( r_t \hat{A}_t,  \text{clip}(r_t, 1-\epsilon, 1+\epsilon)  \hat{A}_t \Big) \Big],  
\end{equation}

where $r_t = \frac{\pi_\theta(a_t|s_t)}{\pi_{\theta_{\text{old}}}(a_t|s_t)}$ is likelihood ratio 
evaluated at the sampled state–action pair $(s_t, a_t)$, $\hat{A}_t = \hat{A}^{\pi_{\theta_{\text{old}}}}(s_t, a_t)$ is the estimated advantage at timestep $t$ (See appendix about more discussions about $\hat{A}$), $\pi_{\theta_{\text{old}}}$ is the policy before the update, and $\pi_\theta$ is the current (updated) policy. The first term inside the $\min$ function is the same term as in the TRPO objective (See Appendix B for a formal definition of the TRPO objective Eq.~\ref{eq:trpo_objective}). It represents the standard policy gradient contribution. The second term  involves a clipping function

\begin{equation} \label{eq:clip}
\text{clip}(r, 1-\epsilon, 1+\epsilon) =
\begin{cases} 
1-\epsilon & \text{if } r < 1-\epsilon, \\[2mm]
r & \text{if } 1-\epsilon \le r_t \le 1+\epsilon, \\[1mm]
1+\epsilon & \text{if } r > 1+\epsilon.
\end{cases}
\end{equation}

$\epsilon$ is a hyperparameter which roughly says how far away the new policy is allowed to go from the old.  When the advantage estimate satisfies $\hat{A}_t> 0$, the objective is increased by enlarging $r_t$. But $r_t$ is clipped at an upper bound of $1 + \epsilon$, and the lower bound is left unconstrained, as gradient ascent naturally drives $\pi_\theta$ upward. Conversely, when $\hat{A}_t < 0$, the objective benefits from reducing $r_t$. In this case, $r_t$ is clipped at a lower bound of $1 - \epsilon$  while the upper bound remains unrestricted for the same reason.

\begin{figure}[htbp]
    \centering
    \begin{subfigure}[b]{0.48\textwidth}
        \centering
        \includegraphics[width=\textwidth]{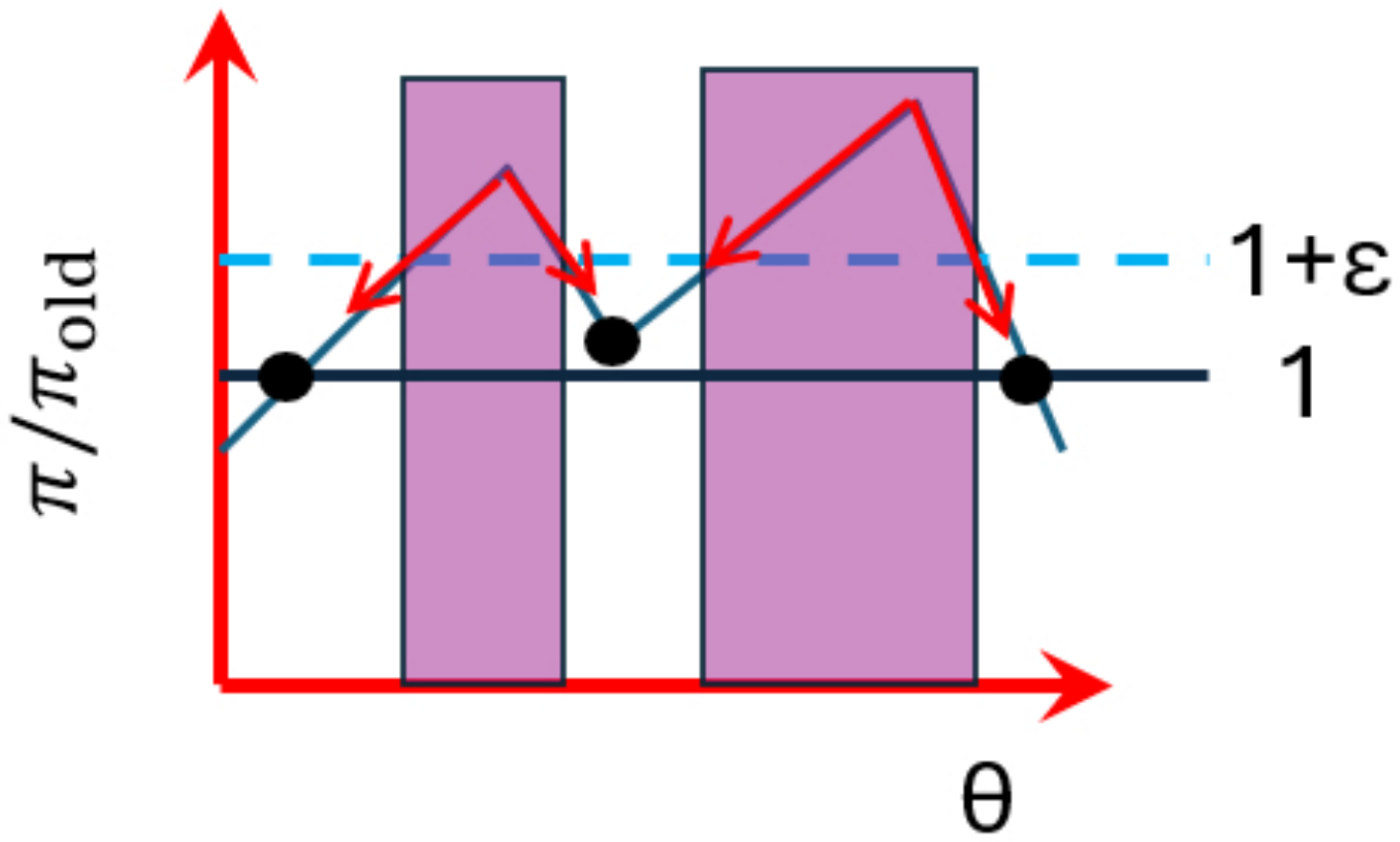}
    \end{subfigure}
    \caption{Clip function illustrated}
    \label{fig:clip_illustrated}
\end{figure}

In PPO, the total loss is a combination of the policy surrogate loss, the critic loss, and an entropy bonus to encourage exploration:

\begin{equation}\label{eq:total_ppo_loss}
L(\theta, \phi) = \underbrace{- \mathcal{J}^{\text{CLIP}}(\theta)}_{\text{policy loss}} 
+ c_1 \underbrace{L^{\text{critic}}(\phi)}_{\text{critic loss}} 
- c_2 \underbrace{L^S(\theta)}_{\text{entropy bonus}}
\end{equation}

Here $L^{\text{critic}}(\phi)$ is the critic loss. It can be the standard loss in Eq.~\ref{eq:ppo_value_loss} or the clipped value loss in Eq.~\ref{eq:ppo_value_loss_clipped}, $L^S$ is the entropy bonus term defined as follows:

\begin{equation}\label{eq:entropy_bonus}
  L^S(\theta) = \mathbb{E}_t \big[ \mathcal{H}(\pi_\theta(\cdot|s_t)) \big]
\end{equation}

where the entropy $\mathcal{H}$ measures how uncertain the policy is at a given state:
\begin{equation}
\mathcal{H}(\pi_\theta(\cdot|s)) = - \sum_{a} \pi_\theta(a|s) \log \pi_\theta(a|s)
\end{equation}

If the policy is deterministic, $\mathcal{H} = 0$. If the policy is uniform/random, $\mathcal{H}$ is maximal. The parameter $c_2>0$ controls how much we encourage exploration. The negative sign indicates that we maximize entropy while gradient descent minimizes the total loss. 

In the standard PPO implementations, an explicit KL-divergence penalty is typically not included. However, a variant of PPO incorporates a KL penalty term directly into the objective to explicitly control the divergence of the updated policy from a reference policy~\cite{schulman2017proximalpolicyoptimizationalgorithms}. This KL-penalized PPO objective can be written as:

\begin{equation} \label{eq:ppo_kl_penalty}
\mathcal{J}^{\text{KL}}(\theta) = \mathbb{E}_t \Big[
\frac{\pi_\theta(a_t|s_t)}{\pi_{\theta_{\text{old}}}(a_t|s_t)} \hat{A}_t
\Big] - \beta D_{\text{KL}}\big[\pi_\theta | \pi_{\theta_{\text{ref}}} \big],
\end{equation}

where $\beta$  controls the strength of the penalty, discouraging the updated policy from deviating too far from the reference policy, which is typically a pretrained LLM. 
In the original PPO paper~\cite{schulman2017proximalpolicyoptimizationalgorithms}, the authors compared  the clipped objective in Eq.~\ref{eq:ppo_objective} and the KL-penalized objective. Their study show that both the clipping and KL-penalized objectives have their merits, but the clipped objective in Eq.~\ref{eq:ppo_objective} generally performed better across a range of tasks and hyperparameters. As a result, the clipped surrogate objective has become the standard choice for PPO implementations. In our experiments with PPO, we do not use this KL-penalized objective.

\paragraph{Observations about the clipping function} The clipping function Eq.~\ref{eq:clip} is intended to constrain the likelihood ratio $r_t$ in the range so that new policy does not deviate from the old policy too much to keep the training stable. However, as shown in Eq.~\ref{eq:ppo_objective}, the clip function only restrains the impact of an overly aggressive new policy on the estimate of the objective. It cannot fundamentally guarantee a strict trust region. 

In Figure~\ref{fig:clip_illustrated} we illustrate how the clip function works and explain why it cannot garantees the TRPO constraint. For simplicity we only consider the case that the advantage estimate is positive. The same logic applies when $r$ falls below the lower clipping range of $1-\epsilon$. Once the likelihood ratio exceeds the higher clipping range $1+\epsilon$, the clipped term becomes constant at $1+\epsilon$.  The objective function Eq.~\ref{eq:ppo_objective} is the clipped function over $\theta$. For the clipped objective, any $\theta$ in the purple region will give the same objective value because the clipped term is constant. If the optimal $\theta$ is in the purple region, it still violates the TRPO constraint even though the likelihood ratio is clipped in the objective function. Other researchers \cite{ilyas2018deep} also found that PPO is fundamentally unable to enforce a strict trust region. It cannot bound the likelihood ratio within the clipping range as it attempts to do~\cite{wang2020truly}. To keep the new policy within a trust region (the horizontal solid black lines in Figure~\ref{fig:clip_illustrated}), we can add the KL constraint into the objective function or the loss function. The goal is to put the optimal $\theta$ toward the black dots.

As pointed out in \cite{ilyas2018deep}, PPO succeeds not because its clipping mechanism strictly enforces the trust region, but because the algorithm leverages sufficient statistical signal in the gradients, benefits from auxiliary stabilization techniques, and the general robustness of stochastic gradient ascent in deep learning landscapes.

\paragraph{PPO in LLM training} In the context of LLM training, the standard reinforcement learning formulation is adapted to the sequence generation setting. Rather than traditional $(\text{state}, \text{action})$ pairs, we treat the generation process in terms of conditional token probabilities $(o_t \mid q, o_{<t})$, where $q$ is a prompt sampled from the dataset and $o$ is an output sequence sampled from the current policy. Here, the partial sequence $o_{<t}$ together with the prompt $q$ serves as the state, and the next token $o_t$ is the action. Under this interpretation, the PPO objective for LLMs is given by:

\begin{equation}
\mathcal{J}_{\text{PPO}}(\theta) =
\mathbb{E}_{(q,a) \sim \mathcal{D},\, o_{1:T}  \sim \pi_{\theta_{\text{old}}}(\cdot \mid q)}
\Bigg[
\min \Bigg(
\underbrace{\frac{\pi_{\theta}(o_t \mid q, o_{<t})}{\pi_{\theta_{\text{old}}}(o_t \mid q, o_{<t})} \hat{A}_t}_{\text{unclipped term}},\;\;
\underbrace{\text{clip}\Big(\frac{\pi_{\theta}(o_t \mid q, o_{<t})}{\pi_{\theta_{\text{old}}}(o_t \mid q, o_{<t})}, 1-\epsilon, 1+\epsilon\Big) \hat{A}_t}_{\text{clipped term}}
\Bigg)
\Bigg],
\end{equation}

where $o_t$ is the token generated at step $t$ conditioned on the prompt $q$ and the preceding tokens $o_{<t}$. The expectation is taken over prompts $q$ from the dataset $\mathcal{D}$ and sequences sampled from the old policy $\pi_{\theta_{\text{old}}}$. The goal is to update the policy $\pi_\theta$ to maximize this objective, thereby improving the quality of generated sequences according to the advantage estimates $\hat{A}_t$.

In LLM training, a per-token KL penalty is often added to the reward to keep the updated policy close to a reference model~\cite{ouyang2022traininglanguagemodelsfollow}, typically the original pretrained language model. This prevents the policy from drifting too far and producing incoherent or degenerate text. With a reward model $r_{\psi}$, the modified per-token reward becomes:

\begin{equation}\label{eq:ppo_llm_kl_penalty}
r_t =r_{\psi,t} - \beta 
\log \frac{\pi_\theta(a_t|s_t)}{\pi_{\text{ref}}(a_t|s_t)}
\end{equation}

where $\pi_{\text{ref}}$ is the reference model, $\pi_\theta$ is the updated policy,  and  $\beta$ controls the strength of this penalty. This KL-regularized reward helps prevent the policy from drifting too far from the original language model, maintaining fluency and coherence in generated text. In our PPO experiments, we include this per-token KL penalty in the reward function as shown in Eq.~\ref{eq:ppo_llm_kl_penalty}.

\paragraph{Cautious note about PPO} In the study of \cite{ilyas2018deep} they found that PPO demonstrates several fundamental deviations from its motivating theory. First, although PPO is intended to constratin policy updates via clipping, the algorithm is fundamentally unable to enforce a strict trust region. Second, the objective function that PPO optimizes is often not reflective of the underlying true reward landscape. It is observed freuqently that moving in the directions that increase the surrogate reward using PPO can actually leads to a decrease in the true reward. Third, both TRPO and PPO consistently operate with poor estimate of the true gradient.  Albeit these issues, TRPO and PPO can consistently increase the reward~\cite{ilyas2018deep}.

\subsection{Grouped Relative Policy Optimization}

The GRPO objective extends PPO to the LLM setting, where training is performed over grouped rollouts of generated responses. The algorithm maximizes the following surrogate objective:

\begin{equation}\label{eq:grpo_objective}
\begin{aligned}
\mathcal{J}_{\text{GRPO}}(\theta) 
= \ & \mathbb{E}_{\, q \sim \mathcal{D},\, \{o_i\}_{i=1}^G \sim \pi_{\theta_{\text{old}}}(\cdot \mid q)} \Bigg[
     \frac{1}{G} \sum_{i=1}^G \frac{1}{|o_i|} \sum_{t=1}^{|o_i|}
     \Big(
        \min\big( r_{i,t}(\theta)\hat{A}_{i,t},\ 
                  \text{clip}( r_{i,t}(\theta), 1-\varepsilon, 1+\varepsilon ) \hat{A}_{i,t}
        \big) \\
& \qquad\qquad\qquad\qquad
        -\ \beta\, D_{\text{KL}}(\pi_{\theta} \| \pi_{\text{ref}})
     \Big)
   \Bigg].
\end{aligned}
\end{equation}

where:

\begin{itemize}
    \item $q$: question sampled from the question dataset $\mathcal{D}$.
    \item $\{o_i\}_{i=1}^{G}$: A group of $G$ outputs  sampled from the old policy $\pi_{\theta_{\text{old}}}$ given the same question $q$.
    \item $o_i$: The $i$-th generated output in the group for question $q$.
    \item $|o_i|$: Length of the $i$-th generated output.
    \item $t$: Token index within a generated output.
    \item $r_{i,t}(\theta) = \frac{\pi_\theta(o_{i,t} | q,o_{i,<t})}{\pi_{\theta_{\text{old}}}(o_{i,t} | q,o_{i,<t})}$: Probability ratio between the new and old policies at token $t$.
    \item $\hat{A}_{i,t}$: Estimated advantage at token $t$ in the $i-$th output. The advantage is computed using group-relative normalization, as described below in Eq. (\ref{eq:grpo_advantage}).
    \item $\text{clip}(r_{i,t}(\theta), 1-\varepsilon, 1+\varepsilon)$: PPO-style clipping function.
    \item $D_{\text{KL}}(\pi_\theta | \pi_{\text{ref}})$: KL divergence between the current policy and a reference model.
    \item $\beta$: Coefficient controlling the strength of the KL penalty term.
\end{itemize}

\paragraph{Group-Relative Advantage Estimation}

The foundational principle behind modern policy gradient methods, including REINFORCE and its extensions, is the reduction of variance in gradient estimation through the use of a baseline. Theory posits the optimal baseline as the state value function, $V(s)$, leading to the actor-critic framework, of which PPO is a stable and prominent example. However, GRPO introduces a novel approach to baseline and advantage estimation that sidesteps the complexity and potential instability of training a dedicated critic. Instead of subtracting a learned value function, GRPO computes a relative advantage within a batch of outputs generated by the old policy for a given prompt, $q$. Specifically, for a group of outputs $\{o_i\}_{i=1}^{G}$ with corresponding rewards $\vec{r} = \{r_1, \dots, r_G\}$, the advantage $\hat{A}_{i,t}$ for output $o_i$ is calculated relative to the group's statistics, as defined by:
\begin{equation}\label{eq:grpo_advantage}
  \hat{A}_{i,t} = \frac{r_i - \mathrm{mean}(\vec{r})}{\mathrm{std}(\vec{r}) + \epsilon}, \quad \vec{r} = \{r_1, \dots, r_G\}
\end{equation}

The $\epsilon$ is a small positive number to prevent the denominator from being zero. This method effectively normalizes the reward within the local context of the sampled group, acting as a dynamic, sample-based baseline that reduces variance and facilitates comparison among concurrently generated responses without relying on an external, learned value function. One cautious note is when all rewards in the group are identical or nearly identical, the variance is a zero or near zero, which can lead to disproportionately large advantage estimates. One way is to add more responses if the initial responses have the same reward~\cite{yu2025dapo}.

GRPO normalizes the rewards across a group of outputs from a prompt.  Hu et al. normalize the reward over the mini-batch that includes multiple prompts~\cite{hu2025reinforce++}. A recent study shows that the batch normalization approach shows better performance than the group normalization~\cite{liu2025understanding}.

The number of outputs $G$ in the group is a hyperparameter that can be tuned based on computational resources and desired variance reduction.  Wu et al.~\cite{wu2025takes} demonstrated that a minimum group size of $G=2$ is necessary to achieve stable training dynamics. They showed that this stems from a theoretical reinterpretation that identifies GRPO as a form of contrastive learning, establishing a conceptual connection to Direct Preference Optimization (DPO).  Theoretical analysis validates  that 2-GRPO preserves the contrastive objective and provides unbiased gradient estimates. Furthermore, the potential increase in gradient variance due to the small group size is mitigated by compensating with a larger number of prompts in the mini-batch. 

If we compare the surrogate objective of GRPO (Eq.~\ref{eq:grpo_objective}) with that of PPO (Eq.~\ref{eq:ppo_objective}), we observe that GRPO includes the KL penalty term explicitly in the objective, while PPO typically does not. The KL penalty in GRPO serves to regularize the policy update in case there is a large variance in the advantage from a small $G=2$.

Note here, GRPO uses a coarse-grained credit assignment paradigm: all tokens within a response share the same reward based on the final outcome, each token assigns a uniform reward inversely propotional to the response length. This can be a problem particularly in long-chain reasoning tasks ~\cite{tan2025gtpo}.  Also, the credit assignment favors shorter responses. Several approaches have been proposed to address this issue. Guo et al. propose a method where intermediate reasoning steps receive rewards, enhancing learning for complex tasks~\cite{guo2025deepseek}. Tan et al. introduce GTPO, which incorporates a learned value function to provide token-level advantage estimates, improving credit assignment in lengthy reasoning chains~\cite{tan2025gtpo}.

\paragraph{Clipping and KL Regularization}

GRPO stabilizes policy updates through clipping and KL regularization. The clipping parameter $\varepsilon$ restricts the range of policy updates, while the KL penalty, scaled by $\beta$, prevents excessive divergence from a reference policy $\pi_{\text{ref}}$.  The KL divergence  used in LLM training is aften different from the exact forward KL used in Eq.~\ref{eq:grpo_objective}. Many implementations use an unbiased estimator introduced by Schulman et al. ~\cite{schulman2020approximating}, which takes  the form:

\begin{equation}
D_{\text{KL}}(\pi_\theta \| \pi_{\text{ref}}) = \frac{\pi_{\text{ref}}(o_{i,t} \mid q, o_{i,<t})}{\pi_\theta(o_{i,t} \mid q, o_{i,<t})} - \log \frac{\pi_{\text{ref}}(o_{i,t} \mid q, o_{i,<t})}{\pi_\theta(o_{i,t} \mid q, o_{i,<t})} - 1
\end{equation}

This estimator is always non-negative and provides a lower-variance proxy for the forward KL.

\paragraph{ Compare PPO and GRPO} Both GRPO and PPO are reinforcement learning algorithms. GRPO can be regarded as a variant of PPO, sharing several core principles. In particular, both algorithms employ a clipped surrogate objective to stabilize training, where the likelihood ratio is constrained. They also rely on advantage estimators to guide policy updates, typically initialize from a supervised fine-tuned (SFT) model, and have demonstrated strong efficacy in enhancing mathematical reasoning in LLMs. 

Despite these similarities, GRPO introduces several distinctive departures from PPO, as summarized below:

\begin{enumerate}
    \item \textbf{Critic and advantage estimation:} 
    PPO is an actor–critic method that requires training a separate value function to support advantage estimation via generalized advantage estimation. In contrast, GRPO eliminates the critic entirely and instead computes advantages through group-relative normalization (Eq.~\ref{eq:grpo_advantage}), which aligns naturally with reward models trained to compare multiple outputs. 
    This design both reduces resource requirements and simplifies the training pipeline.
    \item \textbf{KL handling:} KL divergence is not explicitly in the clipped surrogate objective of the standard PPO (Eq.~\ref{eq:ppo_objective}). GRPO applies the penalty explicitly in the objective (Eq.~\ref{eq:grpo_objective}).

    \item \textbf {Entropy bonus:} PPO often incorporates an entropy bonus term in the loss function to encourage exploration (Eq.~\ref{eq:total_ppo_loss}). GRPO does not include an entropy bonus, relying instead on the clipping and KL penalty for stability. It is obsered that GRPO can suffer from entropy collapse. But this can be addressed via dynamic sampling as in DAPO~\cite{yu2025dapo} or adding controllable entropy~\cite{wang2025arbitrary}.
    
    \item \textbf{Objective granularity:} GRPO computes losses at the \emph{sample level}, averaging over tokens within each sequence and then across sequences. PPO, in contrast, operates at the token level. Extensions such as DAPO refine this approach by weighting longer sequences more heavily through token-level normalization.
\end{enumerate}

\subsection{Decoupled Clip and Dynamic sAmpling Policy Optimization}

DAPO is another variant of the PPO method. It has introduced new techniques to address problems in the GRPO method: namely entropy collapse, reward noise and training instability. The DAPO algorithm samples a group of outputs $\{o_i\}_{i=1}^G$ for each question and answer pair $(q,a)$, and optimizes the policy via the following objective:

\begin{equation}\label{eq:dapo_objective}
\mathcal{J}_{\text{DAPO}}(\theta) = \mathbb{E}_{q \sim \mathcal{D}, \{o_i\}_{i=1}^G \sim \pi_{\theta_{\text{old}}}(\cdot | q)} \left[ \frac{1}{\sum_{i=1}^G |o_i|} \sum_{i=1}^G \sum_{t=1}^{|o_i|} \min \left( r_{i,t}(\theta) \hat{A}_{i,t}, \ \text{clip} \left( r_{i,t}(\theta), 1-\varepsilon_{\text{low}}, 1+\varepsilon_{\text{high}} \right) \hat{A}_{i,t} \right) \right]
\end{equation}
The DAPO objective builds on the GRPO formulation in Eq.~\ref{eq:grpo_objective}. Both approaches employ a clipped surrogate objective, both take expectations over questions and groups of responses sampled from a previous policy, and both sum over advantages for individual outputs. However, DAPO introduces several modifications to address shortcomings observed in the standard GRPO framework.

\begin{itemize}

\item {\textbf{Entropy collapse}} One central challenge in reinforcement learning methods such as GRPO is the entropy collapse that arises when prompts yield uniformly optimal responses. In GRPO, suppose all sampled outputs $({o_i})_{i=1}^G$ for a query are equally correct and thus receive identical high rewards. In this case, the relative advantages across the group collapse to zero, leaving no signal for policy updates.

To mitigate this, DAPO introduces a dynamic sampling strategy. The idea is to guarantee that for each query, the sampled responses contain both correct and incorrect outputs, rather than being entirely homogeneous. If an initial group sampled from the old policy $\pi_{\theta_{\text{old}}}$ produces only all-correct or all-incorrect outputs, more samples are drawn from the batch until diversity is achieved or no more samples can be drawn from the batch. By enforcing this balance, dynamic sampling ensures non-zero advantages are always available. 

One cautious note when applying the dynamic sampling strategy is that it can improve policy gradients at the cost of increased computational overhead, since more outputs are sampled. One way to mitigate this overhead is to ensure that the gradient is computed for the mini-batch rather than for each individual prompt. Another approach is to use an entropy bonus to encourage exploration, making the model less likely to produce only fully correct or fully incorrect outputs; we adopt this approach in our experiments.

\item {\textbf{Token-level reward}} Another potential risk of dynamic sampling arises when the policy is updated based on a mixed group of correct and incorrect answers. If correct answers are continuously replaced with inferior ones in the sampled batch, the resulting gradient signal may be misleading, potentially degrading the policy—this is particularly concerning when the model is already strong.

Another limitation of GRPO lies in how its objective aggregates losses. GRPO computes the loss at the sample level: token-level contributions are first averaged within each response ($1/|o_i|$) and then aggregated across the $G$ responses (Eq.~\ref{eq:grpo_objective}). This strategy favors shorter responses because the contribution from each sample to the surrogate objective is normalization by $|o_i|$, which effectively down-weights the contributions from longer outputs.

DAPO addresses this by shifting to a token-level policy gradient loss. In DAPO the contribution from each sample is not affected by its response length. The average is over the total length of all $G$ responses, as shown in Eq.~\ref{eq:dapo_objective}. This change will encourage the model to focus on correct answers instead of gaming for short responses.  This is particularly beneficial for complex reasoning tasks requiring multi-step solutions.  On the flip side, it may encourage verbosity and longer outputs that are not necessarily more accurate.

\item {\textbf{Clip-higher}} DAPO introduces the Clip-higher strategy. Here, the clipping function uses asymmetric thresholds, $\varepsilon_{\text{low}}$ and $\varepsilon_{\text{high}}$, allowing the model to adjust more flexibly. Clip-higher allows the policy to explore more aggressively in the positive direction by setting a higher upper clipping bound. Study by Liu et al.~\cite{liu2025understanding} showed that a conservative clipping bound often hinders the connective tokes such as "therefore", and "but". These tokens often serve as as transition words for new directions in reasoning.  However, the clipping bound may have different effects on models of different sizes. Its effect might be case dependent and requires further investigation. Additionally, DAPO removes the KL penalty term from the objective, further promoting exploration during training. Again, there is a risk that without KL regularization, the policy may drift too far from the reference model, leading to incoherent or irrelevant outputs.

\end{itemize}

\subsection{Comparison of PPO, GRPO, and DAPO}

PPO and its variants GRPO and DAPO, all seek to optimize the policy (LLM parameters, $\theta$) within a trust region. However, they introduce critical structural differences to address the specific challenges of aligning LLMs, particularly around stability and efficient credit assignment. The core distinctions are summarized in Table 1 (Table $\ref{tab:rl_comparison_summary}$).

\paragraph{Critic and Advantage Estimation}

The most significant departure from standard PPO is the mechanism for Advantage Estimation ($\hat{A}_t$).
\begin{itemize}
  \item PPO is an actor-critic method. It requires training a separate value function (critic) to compute $\hat{A}_t$ via GAE.
  \item GRPO and DAPO eliminate the critic entirely. Instead, they compute $\hat{A}_t$ using group-relative normalization against a local batch of generated responses $o_i$for the same prompt $q$ (Equation $\ref{eq:grpo_advantage}$). This simplification reduces the computational load and avoids the instability associated with training a secondary critic network.

\end{itemize}

\paragraph{Loss Aggregation and Reward Hacking}

The granularity at which the loss is aggregated dictates the policy's tendency toward conciseness or verbosity:
\begin{itemize}
  \item PPO uses a per-token loss.
  \item GRPO applies a uniform reward to all tokens within a response, inversely proportional to the response length ($1/|o_i|$). As hypothesized, this structure creates an implicit bias toward shorter responses, which can manifest as reward hacking by achieving the required answer without comprehensive reasoning steps.
  \item DAPO reverts to a more equitable token-level loss aggregation (Equation $\ref{eq:dapo_objective}$). By weighting each token's contribution uniformly across the total length of the generated group, DAPO encourages the policy to generate the necessary lengthy reasoning chains without penalizing response length.

\end{itemize}

\paragraph{Regularization via Clipping and KL Divergence}

All three methods utilize mechanisms to control policy divergence:
\begin{itemize}
  \item PPO employs the standard symmetric clip [$1-\epsilon, 1+\epsilon$] and uses a per-token KL penalty in the reward function to prevent policy drift from the reference model ($\pi_{\text{ref}}$).

  \item GRPO includes the KL penalty explicitly in the loss objective (Equation $\ref{eq:grpo_objective}$), which, as our parametric study shows, must be non-monotonically tuned (coefficient $\beta$) to balance stability and learning efficiency.
  
  \item DAPO omits the KL penalty from the objective.

\end{itemize}

\paragraph{Exploration via Entropy Bonus and Dynamic Sampling}

All three methods utilize mechanisms to control policy divergence:

\begin{itemize}
  \item PPO often incorporates an entropy bonus in the loss function to encourage exploration (Equation $\ref{eq:total_ppo_loss}$). 
  
  \item GRPO does not include an entropy bonus. GRPO can suffer from entropy collapse, where the policy becomes overly deterministic.
  \item DAPO addresses entropy collapse through dynamic sampling, ensuring that each prompt's sampled responses contain both correct and incorrect outputs,  although this strategy is computationally costly and, as shown in our results, did not guarantee improved performance.
    \item DAPO introduces an asymmetric clip [$1-\epsilon_L, 1+\epsilon_H$], allowing for more aggressive exploration of higher-reward actions while constraining downward policy deviation. 
  
\end{itemize}

\begin{table}[h!]
\centering
\caption{Key Observations and Comparison of RL Strategies for LLM Training}
\label{tab:rl_comparison_summary}
\begin{tabularx}{\textwidth}{p{4cm} p{2.2cm} p{2.2cm} p{2.2cm} >{\raggedright\arraybackslash}X}
\toprule
\textbf{Feature} & \textbf{PPO} & \textbf{GRPO} & \textbf{DAPO} & \textbf{Key Finding/Observation} \\
\midrule
\addlinespace
Clipping & $[1-\epsilon, 1+\epsilon]$ & $[1-\epsilon, 1+\epsilon]$ & $[1-\epsilon_L, 1+\epsilon_H]$  &
Can limit the likelihood ratio but cannot enforce strict trust regions. \\
\addlinespace
\addlinespace
\midrule
\addlinespace
Critic/Value Function & Required (Actor-Critic) & Not Required & Not Required & GRPO/DAPO simplify training by eliminating the high-variance critic. \\
\addlinespace
\addlinespace
\midrule
\addlinespace
Advantage Estimation $\hat{A}_t$ & via GAE and Critic & Group-Relative (Mean/Std) & Group-Relative (Mean/Std) & Relative baselines reduce variance without a separate model. \\
\addlinespace
\addlinespace
\midrule
\addlinespace
Loss Aggregation & Token-level & Sample-level & Token-level & GRPO's sample-level loss favors shorter responses; DAPO's token-level loss mitigates this. \\
\addlinespace
\addlinespace
\midrule
\addlinespace
KL Regularization & KL Penalty in Reward & Explicit KL Penalty in Loss & not in standard DAPO & The KL penalty coefficient ($\beta$) in GRPO must be \textbf{non-monotonically tuned} for optimal results. \\

Entropy Bonus & Included & Not Included & Not Included & Entropy bonus in PPO encourages exploration; GRPO can suffer from entropy collapse. \\
\bottomrule \\
\end{tabularx}
\end{table}

\section{Parametric Study}

PPO, GRPO, and DAPO represent three closely related and widely used strategies for training LLMs, all originating from the same foundational principles of PPO but differing in key modifications designed to improve stability, efficiency, or sample quality. Here we conduct a focused parametric study to isolate the influence of several critical hyperparameters. Specifically, we examine the role of the entropy bonus, the impact of group size $G$ in both GRPO and DAPO, KL divergence, the effect of token-level versus sample-level reward formulations, and the contribution of the dynamic sampling strategy introduced in DAPO. Our goal is to provide practitioners with actionable insights into how these design choices affect learning dynamics and downstream performance when training LLMs with reinforcement learning.

\subsection{Entropy bonus}
\begin{figure}[htbp]
    \centering
    
    \begin{subfigure}[b]{0.48\textwidth}
        \centering
        \includegraphics[width=\textwidth]{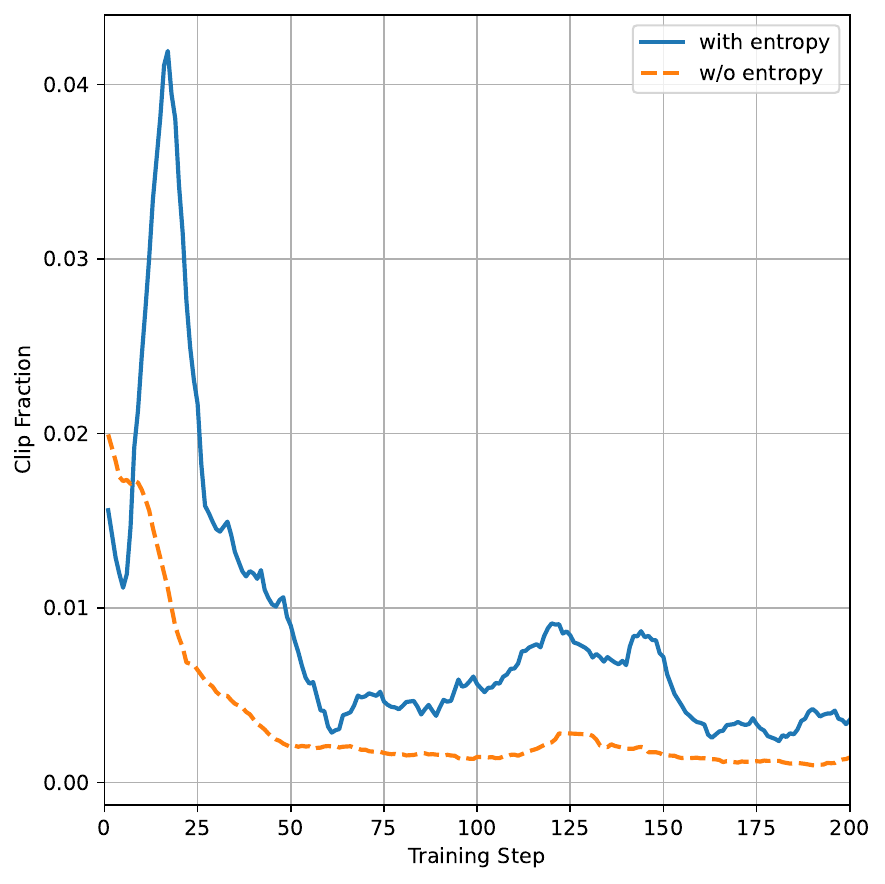}
        \caption{Clip fraction during training} 
        \label{fig:ppo_clipfrac}
    \end{subfigure}
    \hfill
    \begin{subfigure}[b]{0.48\textwidth}
        \centering
        \includegraphics[width=\textwidth]{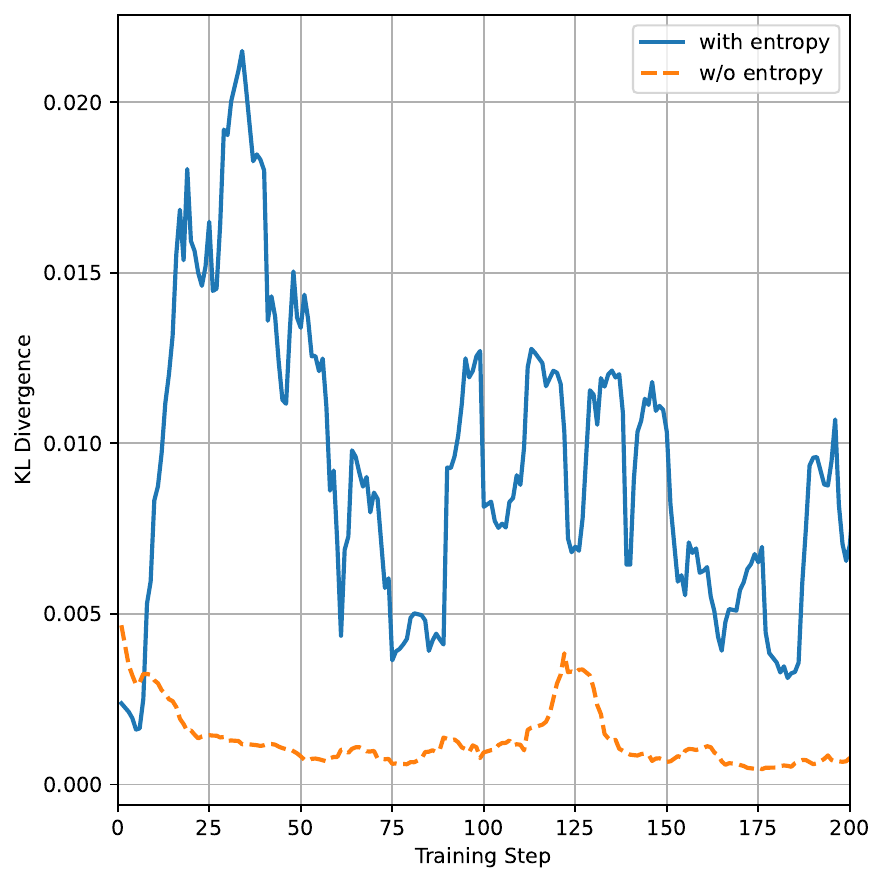}
        \caption{KL divergence between new and old policies} 
        \label{fig:ppo_kl_div} 
    \end{subfigure}

    \vspace{1em} 

    \begin{subfigure}[b]{0.48\textwidth} 
        \centering
        \includegraphics[width=\textwidth]{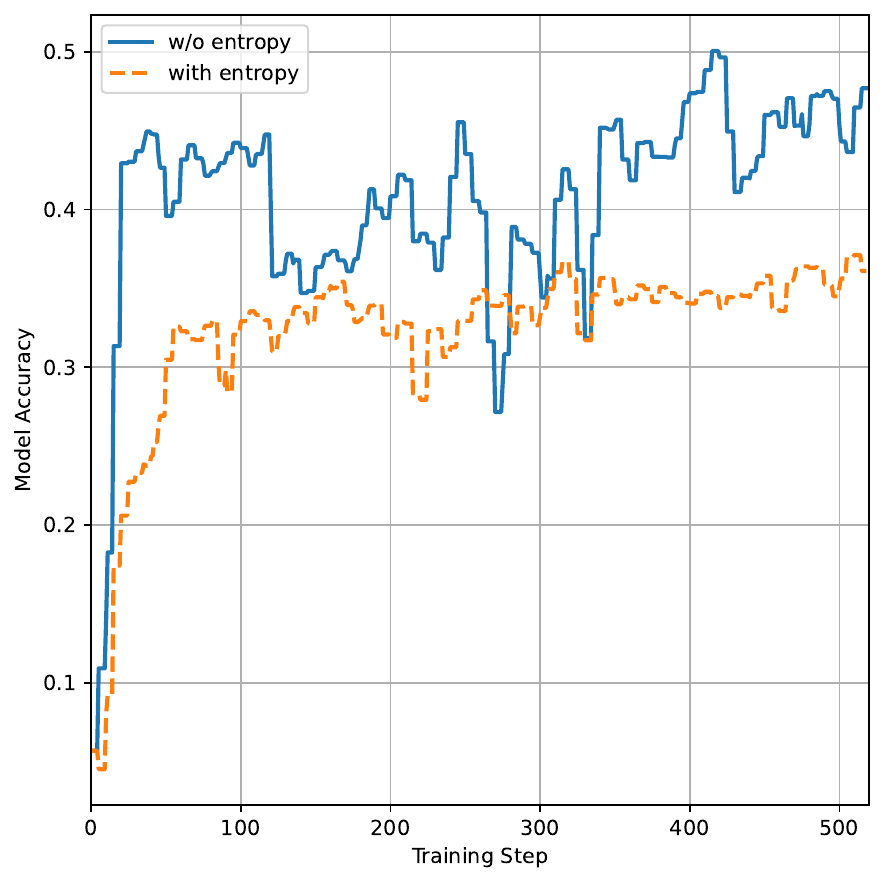}
        \caption{Model accuracy moving average} 
        \label{fig:ppo_model_accuracy_entropy}
    \end{subfigure}
    
    \caption{The effect of entropy bonus on PPO training. (a): The entropy bonus increases the fraction of clipped tokens during training. (b): The entropy bonus enhances the KL divergence between the new and old policies to favor exploration. (c): Model Accuracy performance. Surprisingly, adding an entropy bonus leads to lower accuracy compared to training without entropy regularization.}
    \label{fig:ppo_side_by_side}
\end{figure}

We begin by examining the effect of the entropy bonus coefficient $c_2$  in the PPO loss function (Eq.~\ref{eq:total_ppo_loss}). The entropy term is designed to encourage exploration by preventing the policy from collapsing too quickly into a narrow, deterministic distribution~\cite{schulman2017proximalpolicyoptimizationalgorithms}. A larger $c_2$
 induces more randomness in the policy, while a smaller value allows the model to become increasingly deterministic. As shown in Figure~\ref{fig:ppo_side_by_side}, adding an entropy bonus increases both the fraction of clipped tokens during training and the KL divergence between the updated and reference policies—clear indicators that the policy is exploring more aggressively and undergoing larger updates.

While the entropy bonus promotes exploration, this did not translate to performance gains in our experiments. As shown in $\text{Figure}~\ref{fig:ppo_model_accuracy_entropy}$, the PPO run with an entropy bonus achieved consistently lower model accuracy. This is likely due to the entropy term discouraging exploitation by forcing the model to maintain higher uncertainty, thereby reducing its confidence in optimal actions and slowing the refinement of desired behaviors within the training duration in our experiments (520 steps total).

The original GRPO and DAPO do not include an entropy bonus in their loss functions. The DAPO addresses the entropy collapse issue via dynamic sampling~\cite{yu2025dapo}.

\subsection{Learning Rate on PPO}

\begin{figure}[htbp]
    \centering
    \begin{subfigure}[b]{0.48\textwidth}
        \centering
        \includegraphics[width=\textwidth]{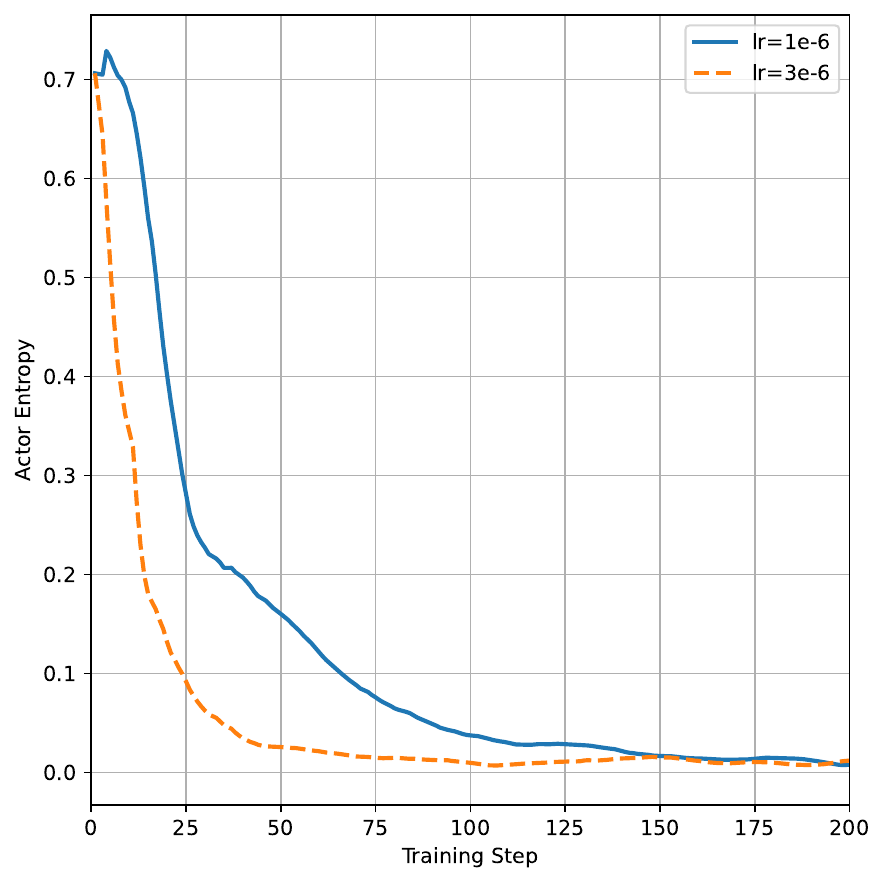}
        \caption{Actor entropy moving average}
        \label{fig:ppo_actor_entropy}
    \end{subfigure}
    \hfill
    \begin{subfigure}[b]{0.48\textwidth}
        \centering
        \includegraphics[width=\textwidth]{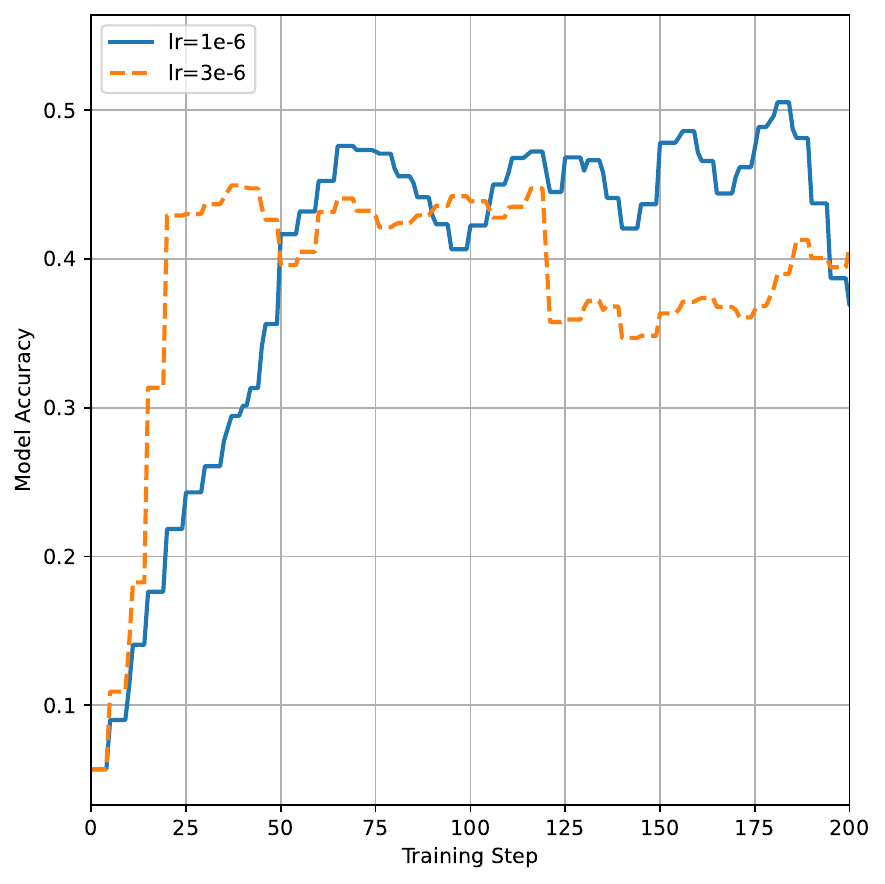}
        \caption{Model accuracy moving average}
        \label{fig:ppo_model_accuracy_lr}
    \end{subfigure}
    \caption{Model performance on GSM8k with different learning rates. A smaller learning rate of \(1 \times 10^{-6}\) leads to more stable training, while a larger learning rate results in higher fluctuations, although the model achieves higher accuracy.}
    \label{fig:ppo_model_lr}
\end{figure}

Figure~\ref{fig:ppo_model_lr} shows how the learning rate affects PPO training. Consistent with prior observations in both PPO~\cite{schulman2017proximalpolicyoptimizationalgorithms} and large-scale RLHF training~\cite{ouyang2022traininglanguagemodelsfollow}, a higher learning rate causes the policy entropy to decay more rapidly (Figure~\ref{fig:ppo_actor_entropy}), indicating that the policy becomes confident and less exploratory much earlier in training.

As shown in Figure~\ref{fig:ppo_model_accuracy_lr}, the higher learning-rate model reaches higher accuracy more quickly, but its trajectory is noticeably less stable and exhibits larger fluctuations. This behavior aligns with previous findings that aggressive optimization steps in PPO can induce training volatility and KL divergence spikes~\cite{schulman2017proximalpolicyoptimizationalgorithms}, and that RLHF pipelines are particularly sensitive to oversized policy updates~\cite{ouyang2022traininglanguagemodelsfollow}. In contrast, the lower learning rate produces slower but more stable improvement, reflecting the common recommendation in RLHF practice to use small learning rates to maintain smooth policy evolution~\cite{ouyang2022traininglanguagemodelsfollow}. This observation is not unique to PPO but generally applicable to policy gradient methods in LLM fine-tuning.

\subsection{GRPO Group Size $G$}

GRPO relies on sampling a group of outputs for each question to compute the relative advantages (Equation~\ref{eq:grpo_advantage}). The group size 
$G$ determines how many responses are sampled per question during training. A larger group provides a more reliable estimate of relative advantages by averaging over a wider range of possible outputs. This reduces the variance of the advantage estimates and stabilizes the policy gradient updates. However, increasing 
$G$ also raises the computational cost, as more responses must be generated and evaluated for each training instance. DeepSeekMath~\cite{shao2024deepseekmathpushinglimitsmathematical} adopted 
$G=64$, a policy learning rate of $1.0\times10^{-6}$, and a KL coefficient of 0.04 to train their 7B model.

\begin{figure}[htbp]
    \centering
    \begin{subfigure}[b]{0.48\textwidth}
        \centering
        \includegraphics[width=\textwidth]
        {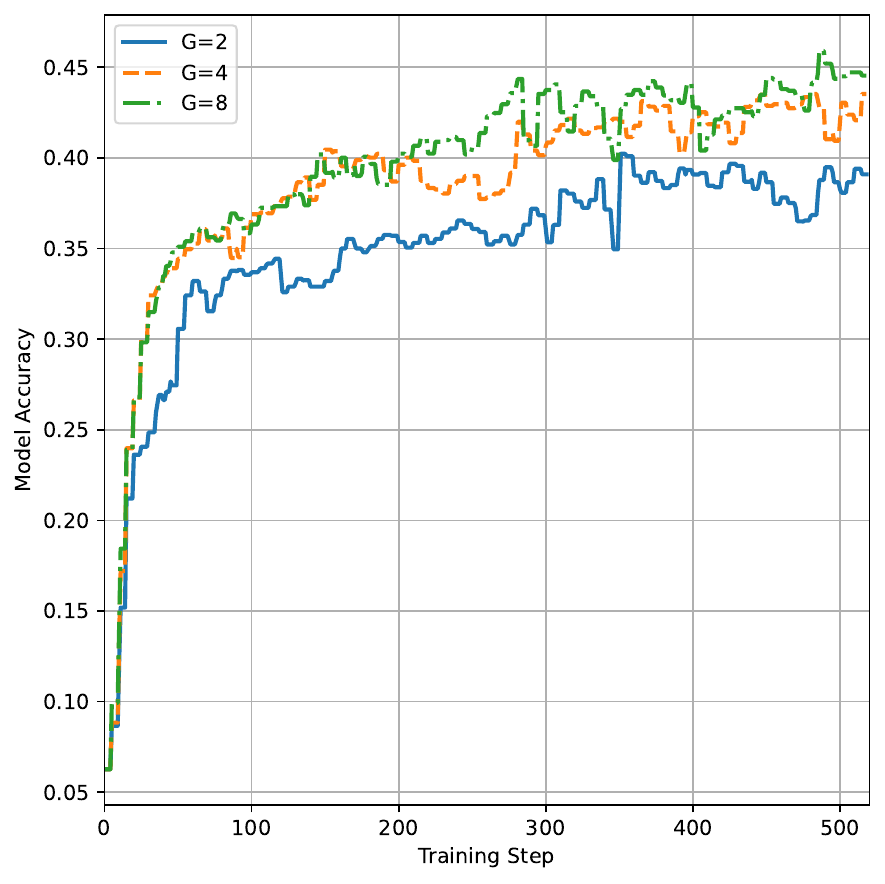}
        \caption{GRPO actor model accuracy moving average}
        \label{fig:grpo_model_accuracy_G}
    \end{subfigure}
    \hfill
    \begin{subfigure}[b]{0.48\textwidth}
        \centering
        \includegraphics[width=\textwidth]{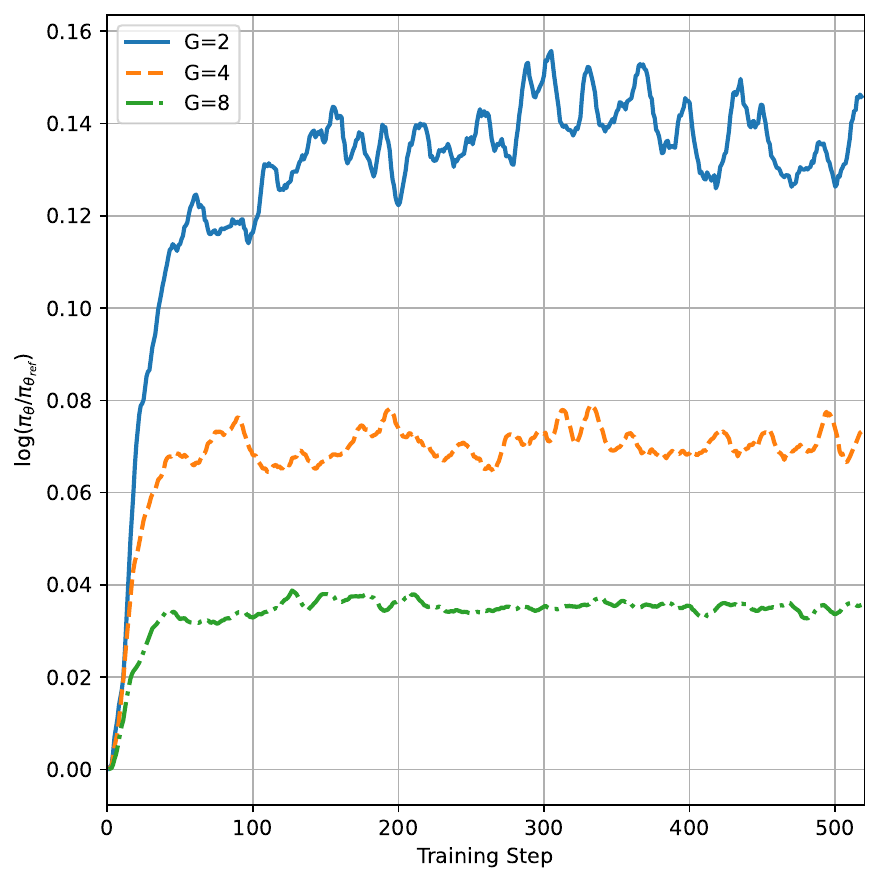}
        \caption{GRPO $D_{\text{KL}}$ moving average}
        \label{fig:grpo_actor_kl_G}
    \end{subfigure}
    \begin{subfigure}[b]{0.48\textwidth}
        \centering
        \includegraphics[width=\textwidth]{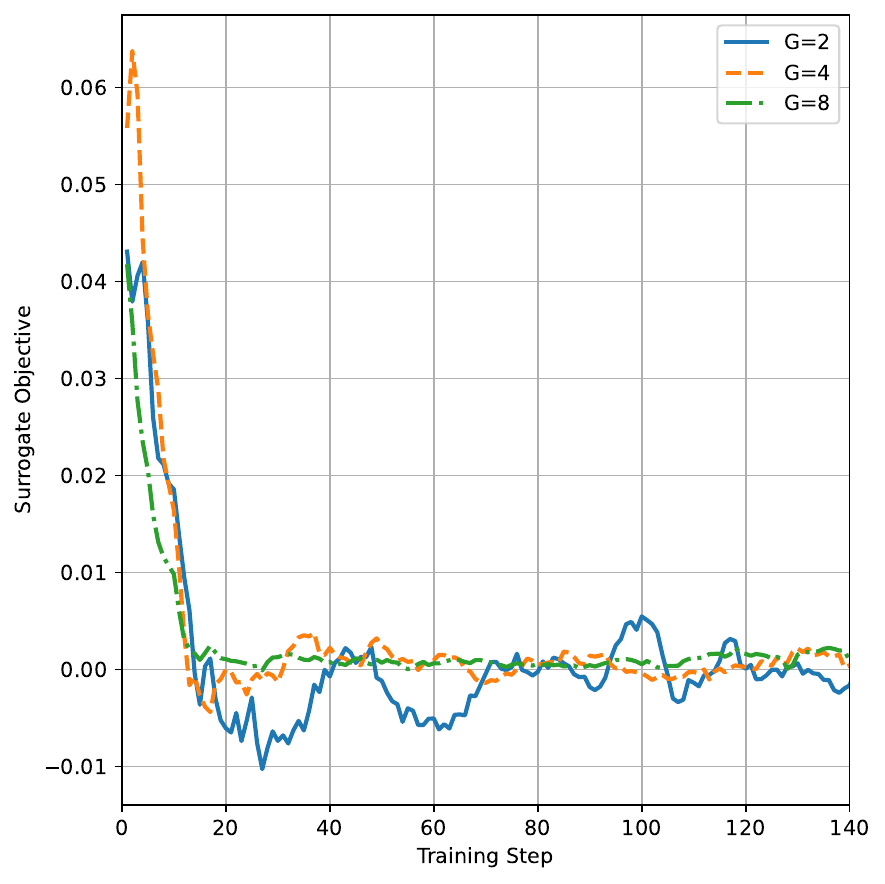}
        \caption{GRPO actor surrogate moving average}
        \label{fig:grpo_actor_surrogate_G}
      \end{subfigure}  
    \caption{GRPO performance on Countdown with different group sizes. A larger group size $G$ leads to better model accuracy and lower KL divergence at higher computation cost. }
    \label{fig:grpo_group_size}
\end{figure}

For our 1.5B model, we experimented with group sizes of $G=2$, $G=4$, and $G=8$. As shown in Figure~\ref{fig:grpo_group_size}, training with 
$G=4$ outperforms $G=2$, while the improvement from $G=4$ to $G=8$ is marginal. Also, a smaller $G$ leads to higher KL divergence between the new and reference policies (Figure~\ref{fig:grpo_actor_kl_G}). As shown in Figure~\ref{fig:grpo_actor_surrogate_G} surrogate objective's volatility decreases with increasing group size ($G$) due to a reduction in variance of advantage estimate ($\hat{A}_{i,t}$). Specifically, the high-frequency fluctuations seen in the $G=2$  run are due to the instability of the $\bar{r}$  and $\sigma_r$  calculated over a minimal sample size. As the group size is increased to $G=4$ and $G=8$, the statistics become less susceptible to individual outliers.  Nevertheless, larger group sizes increase the computational cost per training step. Thus, there is an inherent trade-off between training stability and efficiency.

\begin{figure}[htbp]
    \centering
        \begin{subfigure}[b]{0.48\textwidth}
        \centering
        \includegraphics[width=\textwidth]{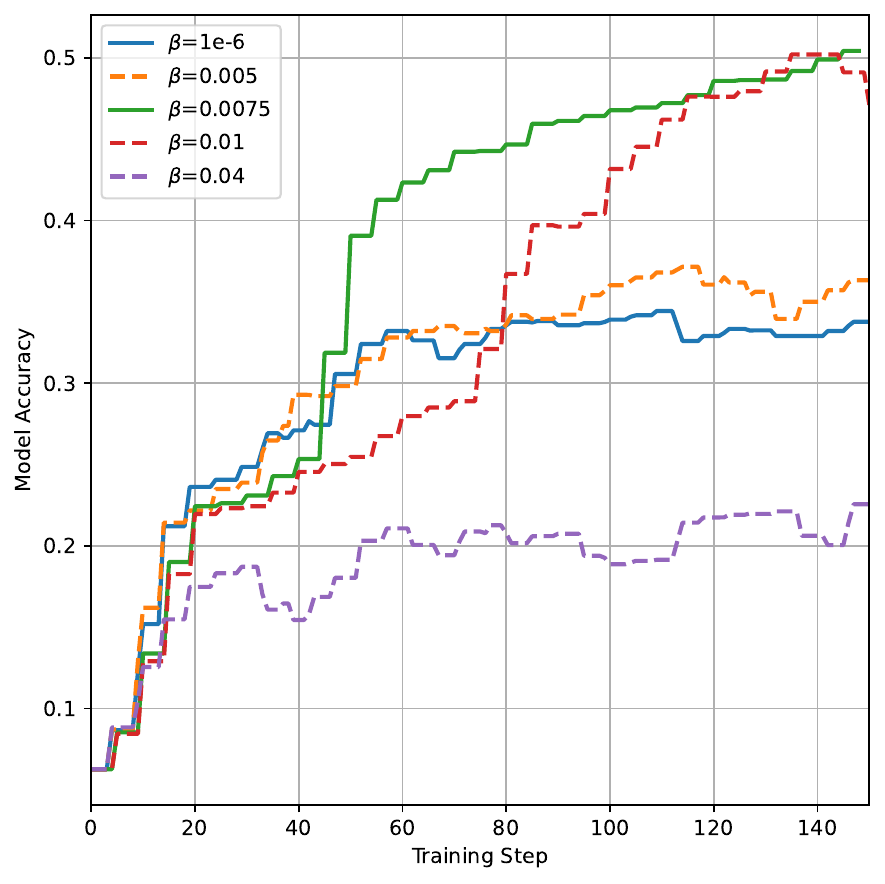}
        \caption{GRPO actor model accuracy moving average}
        \label{fig:grpo_actor_accuracy_G2_beta}
    \end{subfigure}  
    \hfill
    \begin{subfigure}[b]{0.48\textwidth}
        \centering
        \includegraphics[width=\textwidth]
        {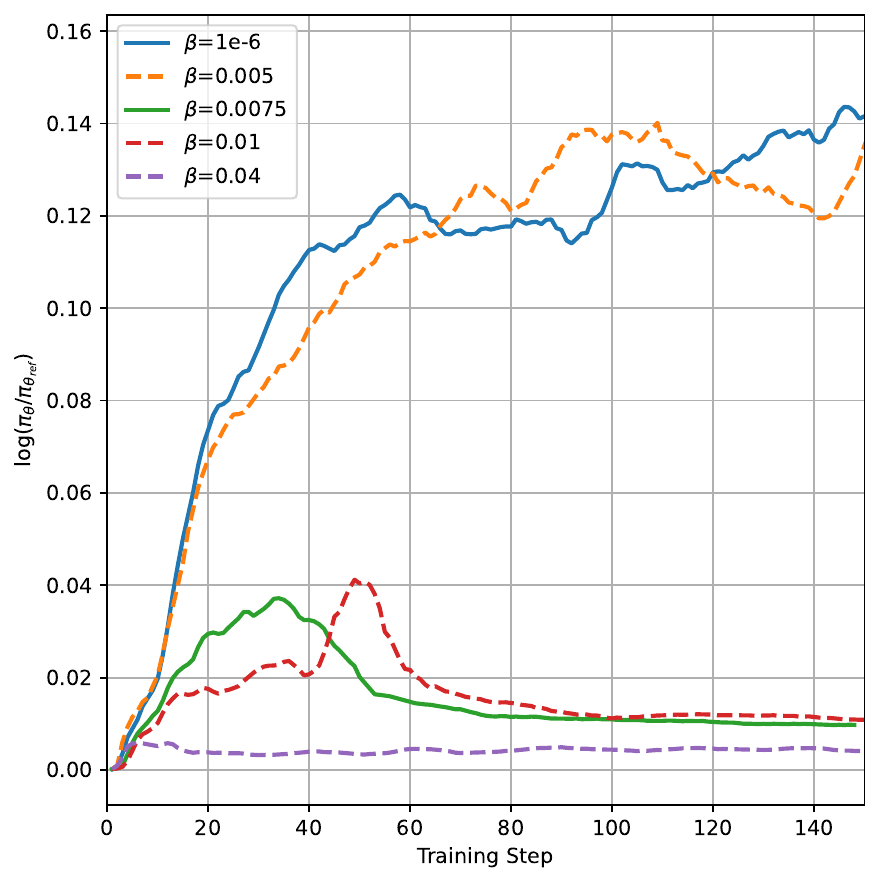}
        \caption{GRPO $D_{\text{KL}}$ moving average}
        \label{fig:grpo_actor_kl_G2_beta}
    \end{subfigure}
  \begin{subfigure}[b]{0.48\textwidth}
        \centering
        \includegraphics[width=\textwidth]{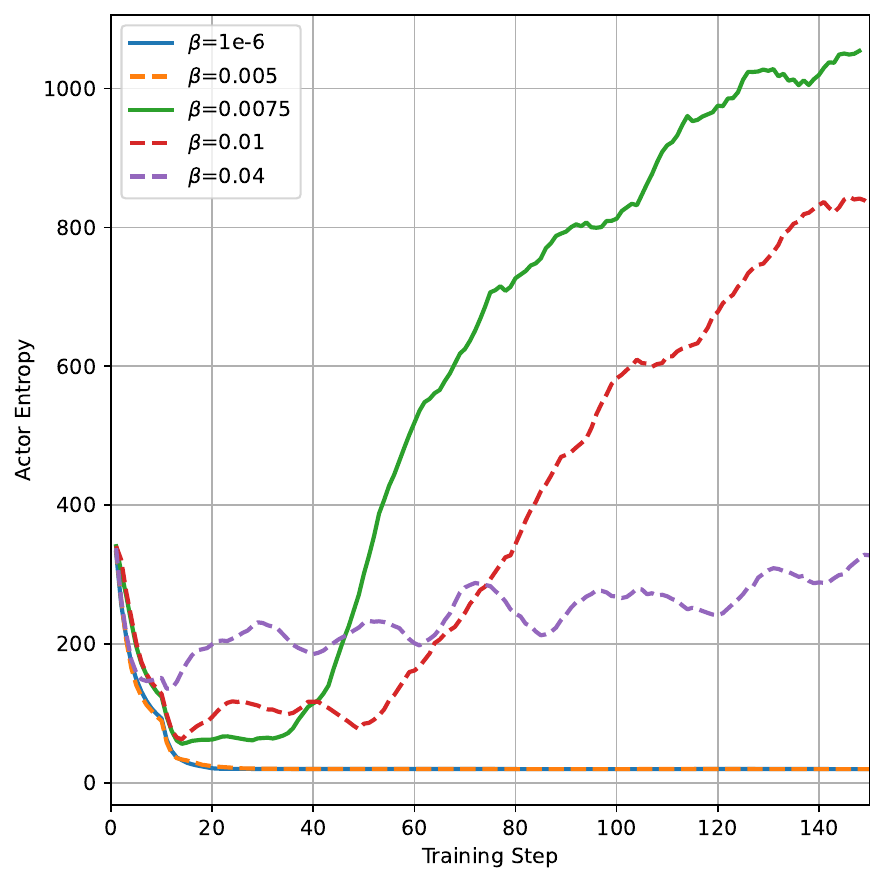}
        \caption{GRPO actor response length average}
        \label{fig:grpo_actor_response_length_G2_beta}  
    \end{subfigure} 
    \hfill
      \begin{subfigure}[b]{0.48\textwidth}
        \centering
        \includegraphics[width=\textwidth]{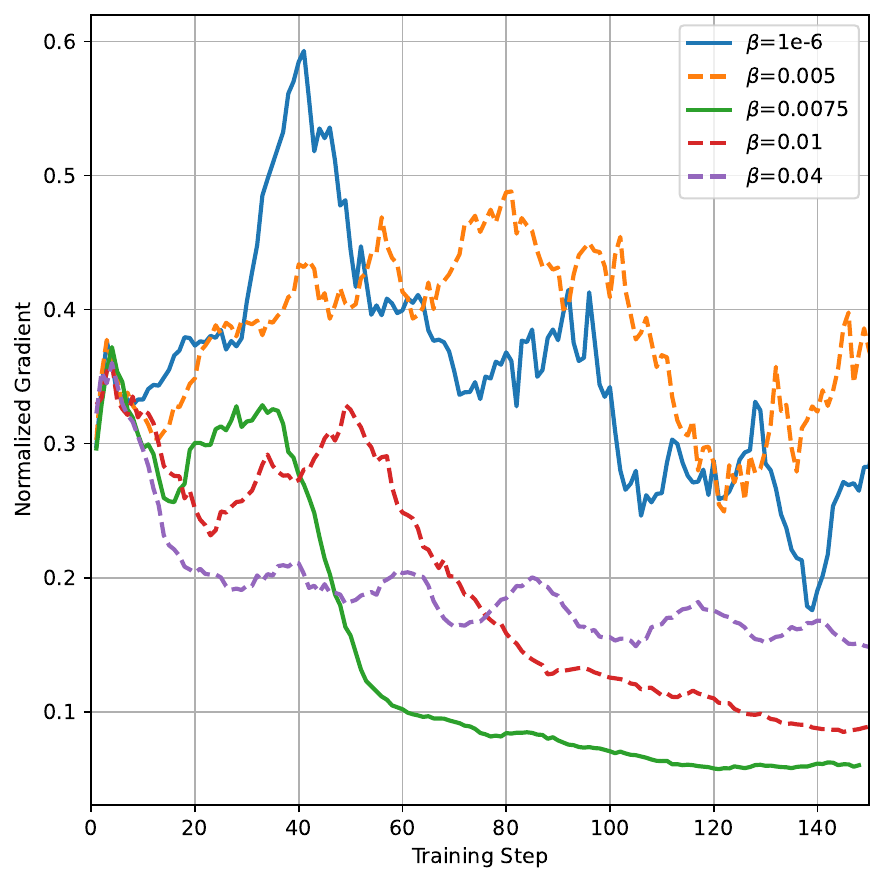}
        \caption{GRPO normalized policy gradient moving average}
        \label{fig:grpo_actor_pg_G2_beta}
    \end{subfigure}  
    \caption{Effect of KL penalty on GRPO performance. The KL penalty coefficient $\beta$ shows  non-monotonic effects on performance.}
\end{figure}

The high variance introduced by small group sizes can be mitigated by increasing the KL penalty coefficient $\beta$  in the GRPO objective (Eq.~\ref{eq:grpo_objective}). A larger $\beta$ constrains policy updates more tightly, limiting deviations from the reference model and thereby improving training stability. Figure~\ref{fig:grpo_actor_kl_G2_beta} illustrates this effect: as $\beta$ increases, the KL divergence between the new and reference policies decreases substantially.

However, the impact of $\beta$ on model accuracy is non-monotonic. As shown in Figure~\ref{fig:grpo_actor_accuracy_G2_beta}, moderate KL coefficients (
$\beta=0.0075$ and $\beta =0.01$) yield the best performance. Further reducing 
$\beta$ does not improve accuracy, while increasing it to $\beta=0.04$ significantly degrades model quality. Excessive regularization constrains the policy too tightly, preventing beneficial exploration and limiting learning efficiency.

Finally, Figure~\ref{fig:grpo_actor_pg_G2_beta} shows that smaller KL coefficients correspond to larger policy gradient magnitudes. However, the increased gradient norm does not translate into better performance. When the KL constraint is weak, the policy diverges more aggressively from the reference, amplifying the variance in advantage estimates and leading to unstable updates. This highlights the importance of balancing the KL regularization strength to achieve both stability and effective learning.

\subsection{Token-level vs Sample-level Loss}

As discussed earlier, DAPO uses a token-level optimization objective (Eq.~\ref{eq:dapo_objective}), whereas GRPO operates at the sample level (Eq.~\ref{eq:grpo_objective}). As a result, GRPO typically produces shorter responses, especially when the KL penalty coefficient is small. In our experiments, the token-level loss in DAPO consistently yielded longer responses, likely because the finer-grained supervision provides feedback at each generation step, encouraging the model to extend its output to accumulate incremental rewards. DRPO exhibits a similar pattern, also generating longer responses than GRPO. However, as we show later, longer responses do not necessarily lead to higher accuracy in the scenarios we examined.

\subsection{Dynamic Sampling Strategy in DAPO}

\begin{figure}[htbp]
    \centering
    \begin{subfigure}[b]{0.44\textwidth}
        \centering
        \includegraphics[width=\textwidth]{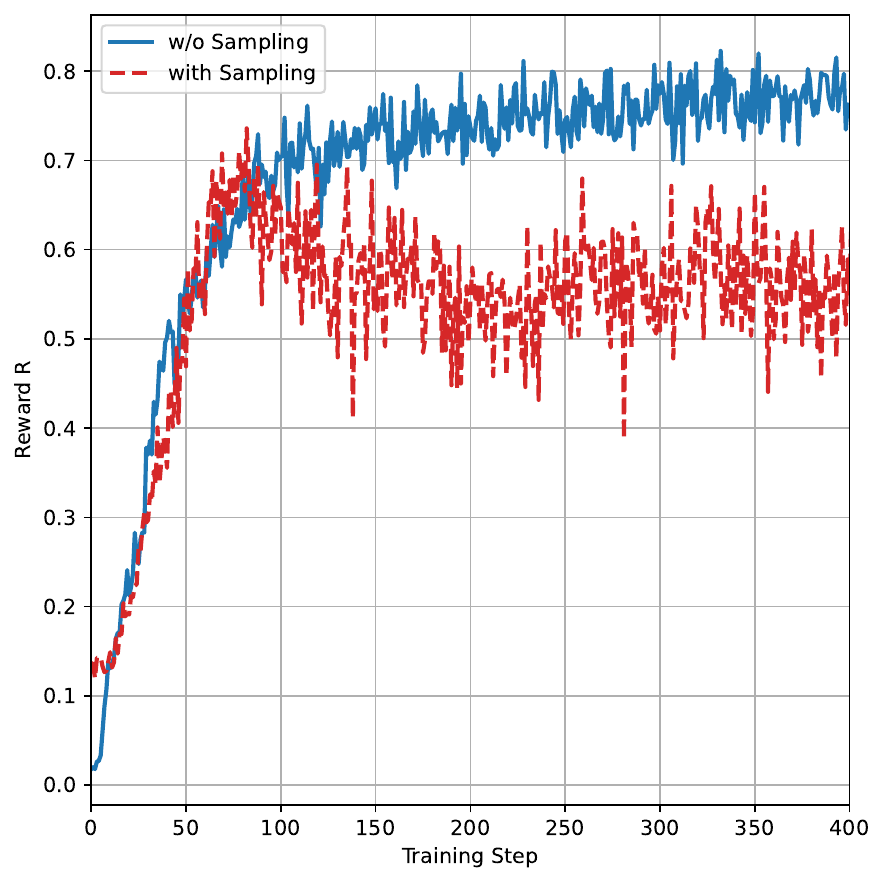}
        \caption{Model accuracy}
        \label{fig:dapo_actor_model_accuracy_filter}
    \end{subfigure}
    \hfill
    \begin{subfigure}[b]{0.44\textwidth}
        \includegraphics[width=\textwidth]{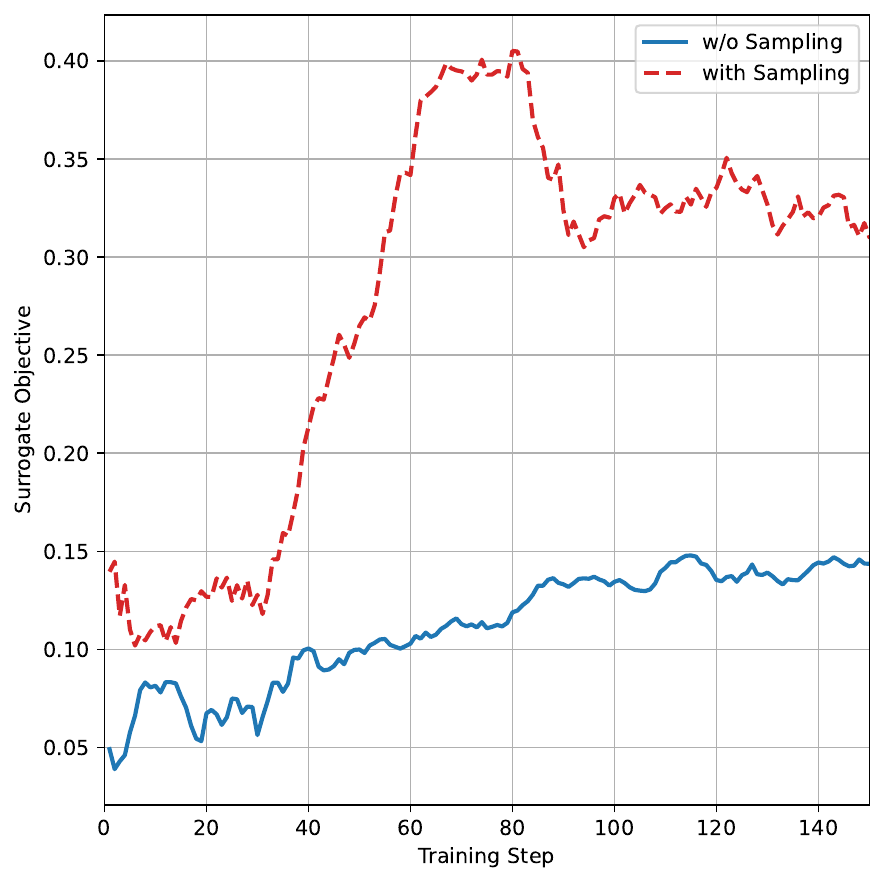}
        \caption{Actor surrogate objective moving average}
        \label{fig:dapo_actor_surrogate_objective_filter}
    \end{subfigure}
      \vspace{1em}
    \begin{subfigure}[b]{0.44\textwidth}
        \centering
        \includegraphics[width=\textwidth]{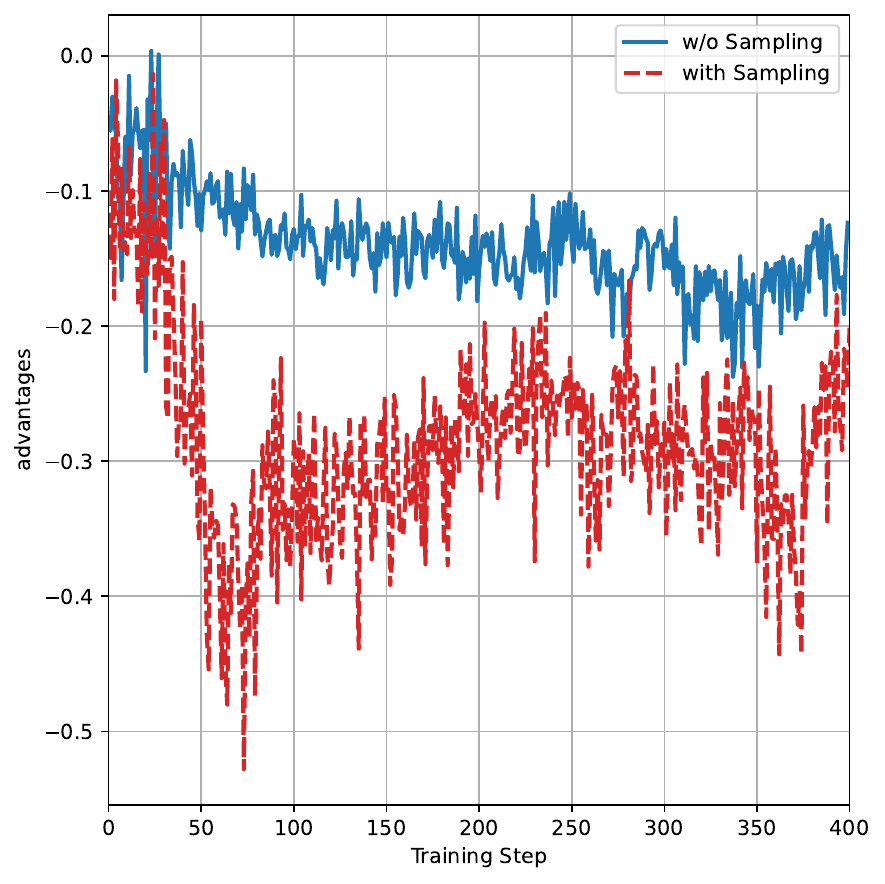}
        \caption{Advantage}
        \label{fig:dapo_critic_advantage_filter}
    \end{subfigure}
    \hfill
    \begin{subfigure}[b]{0.44\textwidth}
        \centering
        \includegraphics[width=\textwidth]{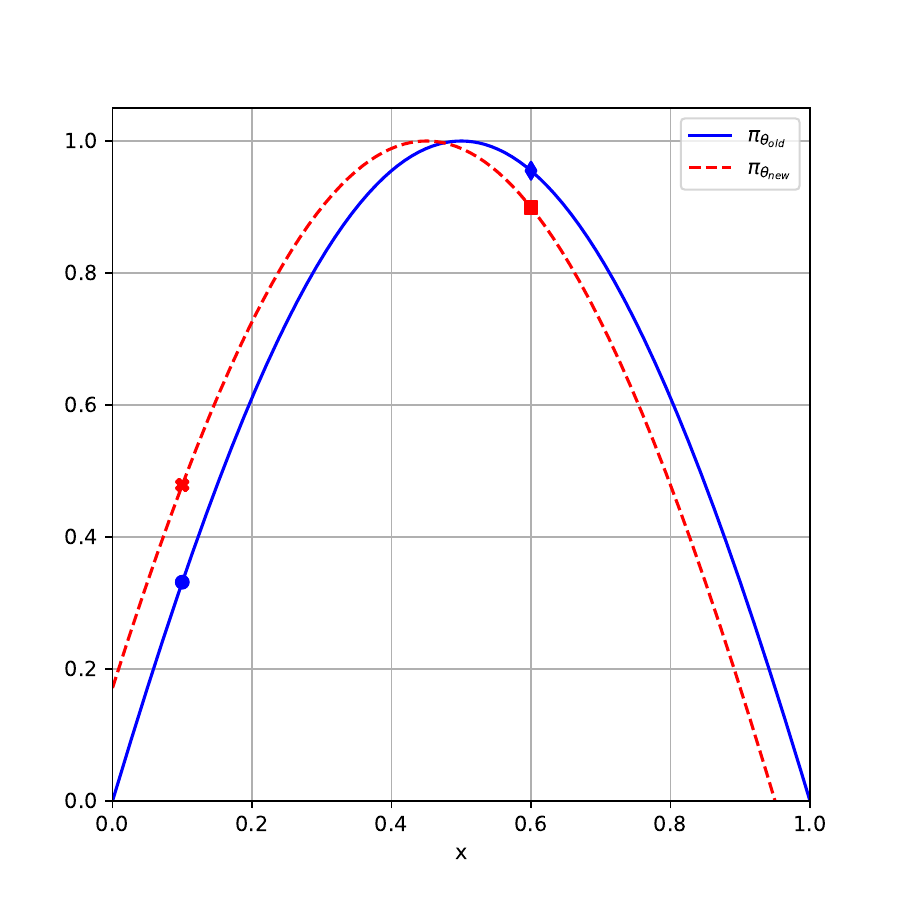}
        \caption{Policy distribution shading}
        \label{fig:dapo_policy_distribution_filter}
    \end{subfigure}

    \caption{Dynamic sampling strategy on DAPO performance. The dynamic sampling strategy improves the surrogate objective and policy gradient, but not necessarily model accuracy.}
    \label{fig:dapo_filter_strategy}  
\end{figure}

Figure~\ref{fig:dapo_filter_strategy} illustrates the impact of the dynamic sampling (DS) strategy on DAPO performance. In these experiments, we fix the generation size to $G=8$. A comparison between the model accuracy (Figure~\ref{fig:dapo_actor_model_accuracy_filter}) and the actor surrogate objective (Figure~\ref{fig:dapo_actor_surrogate_objective_filter} )  reveals a fundamental objective misalignment. While the DS strategy achieves a higher surrogate objective than the training without DS  , this improvement does not translate into a performance gain on the true task-level metric. This suggests that by selectively discarding high-quality samples, the optimization is driven by less informative signals. As shown in Figure~\ref{fig:dapo_actor_model_accuracy_filter}, once the model accuracy peaks around step 75, the DS strategy actually hinders further improvement compared to training without DS.

The underlying issue is that the DS strategy can unintentionally degrade an already strong policy.  To illustrate this phenomenon, consider a policy defined as:

\begin{equation}
    \pi_\theta(x) = \frac{e^{-(x - \theta)^2}-e^{-0.5^2}}{1-e^{-0.5^2}}
\end{equation}

where $\theta$ is the model parameter and $x$ is the prompts or input context, $\pi_\theta(x) $ is the expected probability of the model generating the correct output given the input context $x$.

For a given prompt, for example $x=0.1$, dynamic sampling may remove all correct responses, leaving only inferior ones for advantage computation. In this situation, when a suboptimal response appears slightly better than the group average, it receives a positive advantage. During optimization, the positive signal increases the probability of suboptimal samples, as illustrated by the shift from the blue dot to the red x in Figure~\ref{fig:dapo_policy_distribution_filter}. However, this update can inadvertently reduce the overall policy quality, as the new policy (red square) now have a lower probability of generating the correct response for other prompts compared to the original policy (blue diamond).

This example demonstrates how dynamic sampling can inadvertently reinforce undesirable behaviors: by redefining the comparison set, it biases the gradient updates toward inferior outputs.

In addition, the dynamic sampling strategy substantially increases computational cost. Because the model must generate multiple responses per prompt and discard those that do not meet dynamic sampling criteria, each training step requires more sampling and evaluation. In our experiments, this added over 25\% computation time per training step, with the overhead increasing for longer responses.

Overall, although dynamic sampling can refine the surrogate objective and yield cleaner gradient estimates, it does not guarantee improvements in task-level performance and introduces nontrivial computational trade-offs.

\subsection{Group Size $G$ in DAPO}

\begin{figure}[htbp]
    \centering
    \begin{subfigure}[b]{0.44\textwidth}
        \includegraphics[width=\textwidth]{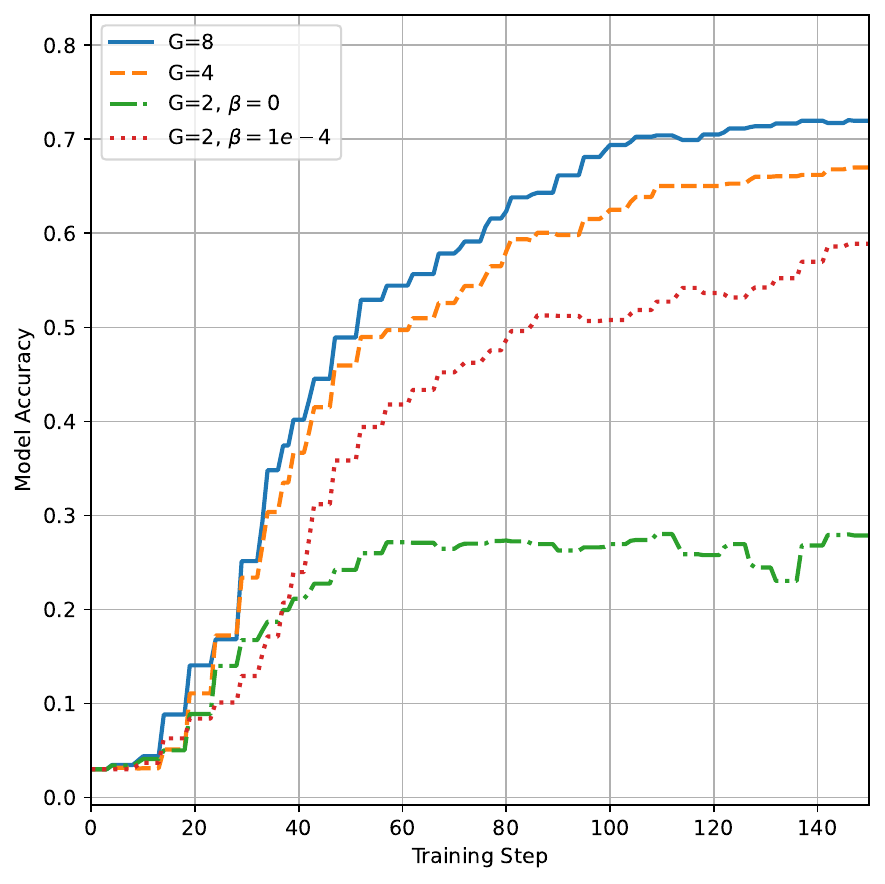}
        \caption{DAPO model accuracy moving average}
        \label{fig:dapo_model_accuracy_G}
    \end{subfigure}
    \hfill
    \begin{subfigure}[b]{0.44\textwidth}
        \centering
        \includegraphics[width=\textwidth]{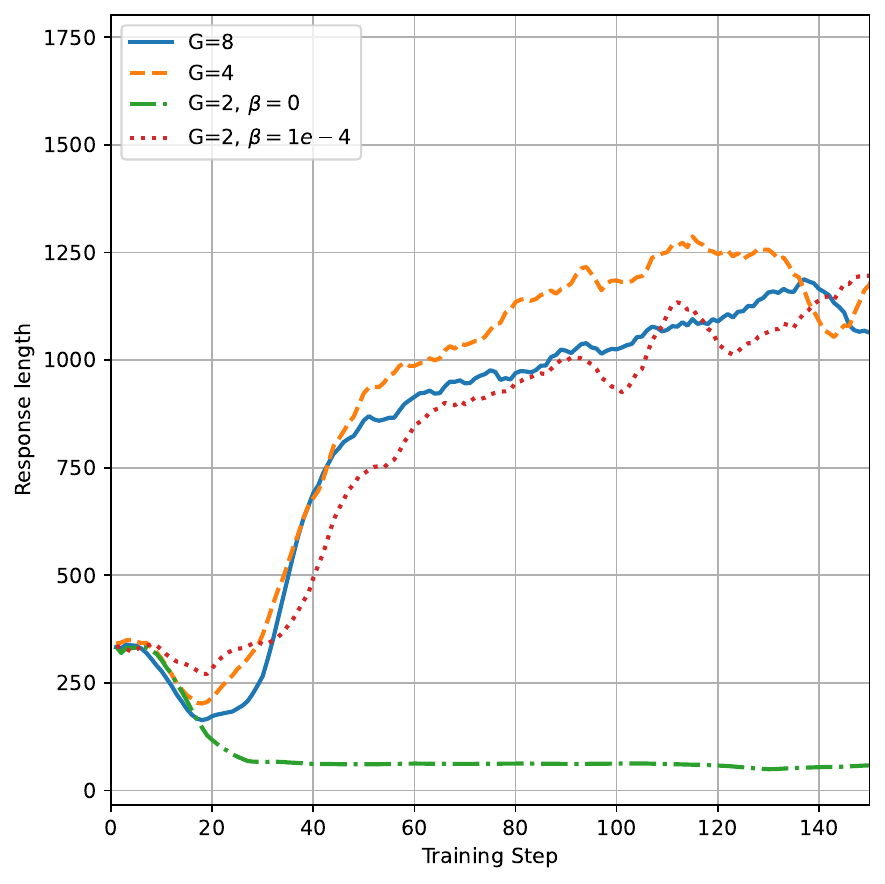}
        \caption{DAPO response length moving average}
        \label{fig:dapo_response_length_G}

    \end{subfigure} 
    \begin{subfigure}[b]{0.44\textwidth}
        \includegraphics[width=\textwidth]{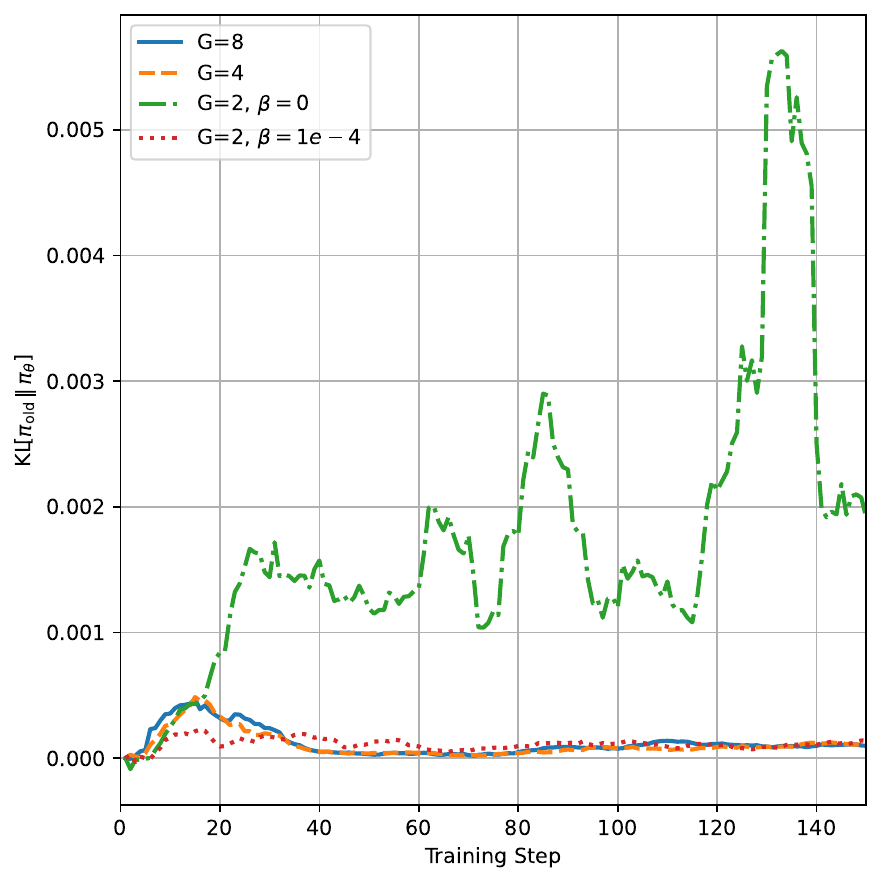}
        \caption{DAPO actor KL divergence moving average}
        \label{fig:dapo_actor_kl_G}
    \end{subfigure}
    \hfill
    \begin{subfigure}[b]{0.44\textwidth}
        \centering
        \includegraphics[width=\textwidth]{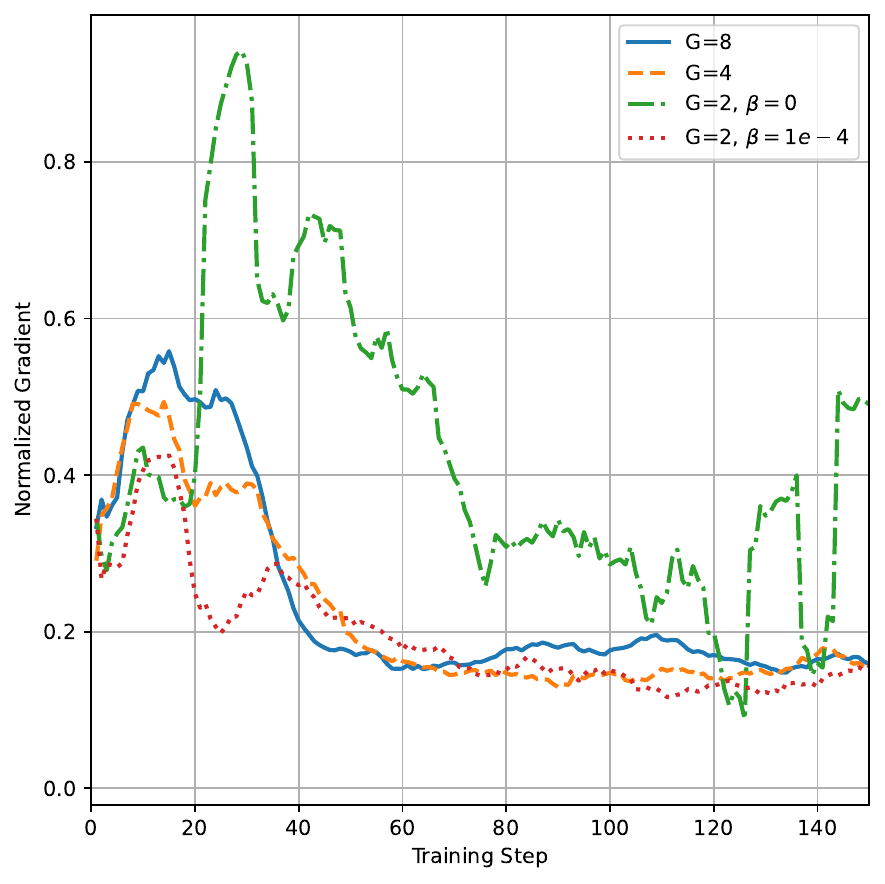}
        \caption{DAPO actor gradient norm moving average}
        \label{fig:dapo_actor_gradient_norm_G}
    \end{subfigure}
    \caption{Effect of group size $G$ on DAPO performance. A larger group size leads to better model accuracy and higher policy gradient.}
    \label{fig:dapo_G}  
\end{figure}

Figure~\ref{fig:dapo_G} illustrates the effect of group size $G$
on DAPO performance. Consistent with GRPO, a larger group size leads to higher model accuracy (Figure~\ref{fig:dapo_model_accuracy_G}). When $G=2$ and the KL divergence penalty is removed, the model performs poorly. Without the KL regularization, the new policy deviates substantially from the old policy (Figure~\ref{fig:dapo_actor_kl_G}), and the generated responses are noticeably shorter compared to those from larger group sizes or runs with KL penalty applied (Figure~\ref{fig:dapo_response_length_G}). The model trained without the KL penalty at $G=2$ also exhibits much larger policy gradient magnitudes, which likely contributes to its training instability.

\subsection{Reward Hacking}

We observed a tendency toward reward hacking in GRPO, particularly when the KL penalty coefficient ($\beta$) was small ($\text{Fig.}~\ref{fig:grpo_actor_response_length_G2_beta}$). As $\beta$ decreased, GRPO generated shorter, less comprehensive responses. This occurs because GRPO's sample-level loss aggregation implicitly favors brevity, leading the model to exploit the reward structure by producing minimal, sometimes reasoning-free, answers that still receive a positive final reward. This tendency is a form of reward hacking, where the model optimizes the proxy metric  without following the desired behavior (comprehensive reasoning steps).

DAPO's token-level loss mitigates this issue by providing finer-grained supervision at each generation step, thus encouraging the model to generate longer, more detailed reasoning chains necessary for complex problem-solving.

We also found that increasing the format reward weight in the reward model can effectively counter this bias. By explicitly rewarding the presence of necessary structure and completeness, the model is incentivized to generate thorough answers instead of taking shortcuts.

\section{Benchmark Results}

To assess the impact of different RL strategies, we systematically compare the performance of our newly optimized models against the base Qwen2.5-1.5B-Instruct model~\cite{qwen2.5} across four  benchmarks. The selection of a relatively small 1.5B parameter model facilitates rapid experimentation and quick turnaround for iterative testing of the RL algorithms. The benchmarks used are: Grade School Math 8K (GSM8K)~\cite{cobbe2021gsm8k}, BIG-Bench Hard (BBH)~\cite{suzgun2022challenging}, Mathematics Aptitude Test of Heuristics (MATH)\cite{hendrycks2021measuringmathematicalproblemsolving}, and More Robust and Challenging Multi-Task Language Understanding Benchmark (MMLU-Pro)~\cite{wang2024mmlupro}.

The main purpose of this evaluation is to ensure a fair comparison of the intrinsic algorithmic contributions within a specific, controlled environment. Evaluation results for LLMs are highly sensitive to protocol choices such as prompt formatting and stopping criteria. A recent study by Meta~\cite{zhao2025mobilellm} trained a series of sub–1B-parameter models, with their largest model, MobileLLM-R1-950M, supporting a maximum sequence length of 32k tokens. We evaluated this model using the llm-evaluation-harness~\cite{eval-harness} and found that, as shown in Table~\ref{tab:token_perf}, increasing the maximum token budget consistently improves performance on both Math500 and GSM8K. This underscores the need to control for such evaluation settings when comparing results across studies.

To maintain this consistency, all models are evaluated on the exact same benchmark splits using the lm-harness evaluation suite. We set the maximum output token length to 2048.

\begin{table}[h!]
\centering
\caption{Performance on Math500 and GSM8K for different max\_tokens\_per\_call values.}
\begin{tabular}{lcc}
\toprule
\textbf{max\_tokens\_per\_call} & \textbf{Math500} & \textbf{GSM8K} \\\midrule
2500 & 52.2 & 60.3 \\
4000 & 60.4 & 63.6 \\
8192 & 66.8 & 67.8 \\
16000 & 73.4 & 66.9 \\
\bottomrule
\end{tabular}

\label{tab:token_perf}
\end{table}

\paragraph{GSM8K}

GSM8K is a high-quality dataset designed to benchmark mathematical reasoning in language models~\cite{cobbe2021gsm8k}. It consists of 8,500 curated grade-school-level math word problems. GSM8K is widely used to evaluate a model’s ability to perform complex problem-solving.

\paragraph{BBH}

BBH is a subset of the Beyond the Imitation Game Benchmark (BIG-Bench)~\cite{suzgun2022challenging}, a suite designed to assess LLMs across reasoning, mathematical problem-solving, commonsense knowledge, coding, and creative tasks. BBH consists of 23 tasks that require advanced multi-step reasoning. These tasks are constructed to challenge models beyond surface-level memorization.

\paragraph{MATH}

The MATH benchmark is a dataset of 12,500 problems sourced from leading U.S. mathematics competitions~\cite{hendrycks2021measuringmathematicalproblemsolving}. MATH emphasizes advanced problem-solving techniques in algebra, calculus, geometry, and statistics. Many of these problems require mathematical heuristics rather than formulaic computation.

\paragraph{MMLU-Pro}
MMLU-Pro is an enhanced version of the MMLU benchmark \cite{hendrycks2021measuringmassivemultitasklanguage}. Unlike the original MMLU, MMLU-Pro shifts the focus toward complex reasoning. An expanded answer choice set (from 4 options to 10) also makes random guessing less effective.

\subsection{Training Setup}

We fine-tune the Qwen2.5-1.5B-Instruct language model~\cite{qwen2.5}. We utilize the Countdown Game dataset\cite{CountdownTasks3to4} as the exclusive source of reward signals. The Countdown Game is a specialized arithmetic task that demands the LLM generate a verifiable sequence of mathematical steps to reach a target number.This structure provides a confined yet challenging environment for testing RL convergence and stability without resource-intensive, open-ended data collection. Model training is based on Verl~\cite{sheng2024hybridflow}.

\subsection{Model Performance}

Table~\ref{tab:model_comparison} summarizes the performance of the base model and the RL fine-tuned models across four benchmarks. All RL-trained policies outperformed the base model all benchmarks, even when trained on a specialized task like the Countdown Game. Among the methods tested, DAPO without dynamic sampling emerged as the most effective algorithm, achieving the highest accuracy on all four benchmarks, including a peak of $\mathbf{53.3\%}$ on GSM8K and $\mathbf{30.0\%}$ on MMLU-Pro. The gains observed on MATH and BBH are marginal. Maybe because these benchmarks cannot be handled by a small LLM like Qwen2.5-1.5B, or, maybe the Countdown Game does not provide sufficiently rich reward signals to improve performance on these tasks.

\begin{table}[ht]
\centering
\caption{Performance comparison of different models on various benchmarks (\%).}
\label{tab:model_comparison}
\begin{tabular}{lcccc}
\toprule
Model & GSM8K & MATH & BBH & MMLU-Pro \\
\midrule
Base & 48.4 & 23.3 & 35.3 & 25.8 \\
PPO  & 50.3 & 25.1 & 36.8 & 27.1 \\
GRPO & 50.8 & 24.7 & 36.9 & 28.2 \\
DAPO (No DS) & 53.3 & 25.4 & 36.9 & 30.0 \\
\bottomrule
\end{tabular}
\end{table}

\section{Conclusion}

We conducted an empirical study comparing three RL algorithms for fine-tuning LLMs on reasoning-intensive tasks. By systematically varying key hyperparameters such as entropy bonus, learning rate, KL penalty coefficients, group size, and loss aggregation methods, we dissected the nuanced effects of each design choice on training stability and downstream performance across multiple benchmarks. Our main findings are summarized as follows:

\begin{itemize} 

\item Entropy bonuses in PPO encourage exploration but do not necessarily improve model performance.

\item A smaller learning rate stabilizes training dynamics in PPO, leading to smoother policy evolution and more consistent accuracy improvements.

\item In both GRPO and DAPO, a larger group size of $G=8$ consistently leads to more stable training and higher final model accuracy compared to smaller groups like $G=2$. 

\item Our tests show that GRPO's sample-level loss aggregation favors shorter, less comprehensive responses. DAPO mitigates this by adopting a token-level loss aggregation, which encourages the generation of longer, more detailed reasoning chains necessary for complex problem-solving.

\item While designed to ensure non-zero advantage signals, in our tests the dynamic sampling strategy in DAPO did not lead to performance gains. 

\item RL can enhance LLM capabilities even when trained on a narrow task like the Countdown Game.

\end{itemize}

We also summarized these key observations in Table~\ref{tab:rl_comparison_summary_fit} for easy reference by future researchers.

\begin{table}[h!]
\centering
\caption{Summary of Key Observations}
\label{tab:rl_comparison_summary_fit}
\begin{tabularx}{\textwidth}{p{3cm} p{2.2cm} p{2.2cm} p{2.2cm} >{\raggedright\arraybackslash}X}
\toprule
\textbf{Feature} & \textbf{PPO} & \textbf{GRPO} & \textbf{DAPO} & \textbf{Key Finding/Observation} \\
\midrule
Entropy Bonus & Yes & N/A  & N/A &  Promotes exploration but not necessarily increases model performance \\
\midrule
\addlinespace
\addlinespace
Learning rate &  &   &  &  should be adjusted to stabilize training and improve convergence. \\
\midrule
\addlinespace
\addlinespace
Group Size ($G$) & N/A & $\ge 2$  & $\ge 2$ & Larger $G$ (\textbf{$G=8$} vs. $G=2$) leads to higher accuracy and lower advantage variance for both GRPO and DAPO. \\
\midrule
\addlinespace
\addlinespace
Loss aggregation & token-level & sample-level  & token-level & Token-level based DAPO has longer responses than GRPO.\\
\midrule
\addlinespace
\addlinespace
Dynamic Sampling & N/A & N/A & Introduced & Does not improve model accuracy  but increased computation time by $\sim 25\%$. \\
\addlinespace
\bottomrule
\end{tabularx}
\end{table}

\bibliographystyle{unsrt}
\bibliography{references}

\appendix

\section{Vanilla policy gradient methods}

Vanilla policy gradient (VPG) methods~\cite{williams1992simple, sutton1999policy} directly optimize the expected cumulative return:  

\begin{align}\label{eq:vpg}
J(\theta) 
&= \int_{\tau \sim \pi_\theta} P(\tau \mid \pi_\theta) \, R(\tau) \, d\tau 
   && \text{(expected return over trajectories)} \nonumber\\
&= \mathbb{E}_{\tau \sim \pi_\theta} \big[ R(\tau) \big] 
   && \text{(expectation over trajectories under $\pi_\theta$)} \nonumber\\
&= \mathbb{E}_{\tau \sim \pi_\theta} \Big[ \sum_{t=0}^{T} r_t \Big] 
   && \text{(sum of rewards along a finite-horizon trajectory of length $T$)}
\end{align}

where:  
\begin{itemize}
    \item $\theta$: tunable parameters of the policy $\pi_\theta$.
    \item $\pi_\theta$: parameterized stochastic policy.
    \item $\tau$: trajectory, i.e., a sequence of states, actions, and rewards: $\tau=(s_0, a_0, r_0, \dots, s_i, a_i, r_i, \dots)$
    \item $P(\tau | \pi_\theta)$: probability of observing trajectory $\tau$ under policy $\pi_\theta$: 
    
    \[P(\tau | \pi_\theta) = \rho(s_0) \prod_{t=0} \pi_\theta(a_t | s_t) P(s_{t+1} | s_t, a_t)\]

    \item $\rho(s_0)$ is the initial state distribution and $P(s_{t+1} | s_t, a_t)$ is the environment dynamics.
    \item $R(\tau)$: cumulative reward along trajectory $\tau$. For a finite-horizon trajectory of length $T$, this is simply the sum of rewards: 
    \[
        R(\tau) = \sum_{t=0}^{T} r_t,
    \]
    while for potentially infinite trajectories, a discount factor $\gamma \in [0,1]$ can be applied:
    \[
        R(\tau) = \sum_{t=0}^{\infty} \gamma^t r_t.
    \]
    \item $r_t$: reward received at time step $t$.
    \item $J(\theta) = \mathbb{E}_{\tau \sim \pi_\theta}[R(\tau)]$: expected return under the current policy $\pi_\theta$.
\end{itemize}

\section{Trust Region Policy Optimization}

One major chanllenge of the VPG type methods is the learning rate for the policy updating. An improper large learning rate can possibly leads to policy collapse or performance degradation. Trust region policy optimization (TRPO)~\cite{schulman2015trust} addresses this by constraining the policy updates so that the new policy does not differ too much from the old one. This is done via a trust region constraint, inspired by classical optimization. The key idea is to maximize the expected improvement in policy improvement while constraining the divergence between the new and old policies.  TRPO aims to maximize the expected policy improvement using a surrogate objective, while imposing a trust-region constraint on the KL divergence between the new and old policies:

\begin{equation}\label{eq:trpo_objective}
\max_\theta \quad \mathbb{E}_t \left[ \frac{\pi_\theta(a_t|s_t)}{\pi_{\theta_{\text{old}}}(a_t|s_t)} \hat{A}_t \right] \quad \text{subject to:} \quad D_{\text{KL}}\big[ \pi_\theta(\cdot|s_t) \| \pi_{\theta_{\text{old}}}(\cdot|s_t) \big] \le \delta
\end{equation}

where:
\begin{itemize}
    \item $\pi_\theta$: new policy parameterized by $\theta$,
    \item $\pi_{\theta_{\text{old}}}$: policy before the update,
    \item $\hat{A}_t$: advantage estimated from the old policy,
    \item $r_t=\frac{\pi_\theta(a_t|s_t)}{\pi_{\theta_{\text{old}}}(a_t|s_t)}$: likelihood ratio between new and old policies,
    \item $\delta$: small positive constant controlling the maximum step size (e.g., 0.01),
\end{itemize}

Instead of directly maximizing the true reward as defined in Eq.~\ref{eq:vpg}, modern policy gradient methods optimize a surrogate objective. As noted in \cite{ilyas2018deep}, estimating the true reward in the low-sample regime typical of practical settings often produces a poorly-behaved landscape, creating barriers to direct reward optimization. By contrast, the surrogate objective provides a smoother and more stable landscape, enabling more reliable gradient-based updates, even though it may only approximate the true reward. However, it is important to note that the surrogate reward might not always serve as a reliable proxy: in both low- and high-sample regimes, increases in the surrogate objective do not necessarily correspond to increases in the true reward~\cite{ilyas2018deep}.

Under certain assumptions, the constrained form of TRPO offers strong theoretical guarantees for monotonic policy improvement. However, it requires solving a constrained optimization problem that
 requires second-order optimization methods like Fisher matrix estimation and conjugate gradient computations, which can be computationally expensive and challenging for very large models or environments with high-variance returns~\cite{schulman2017proximalpolicyoptimizationalgorithms}.

To simplify the optimization, the hard KL-divergence constraint can be transformed into a soft penalty term added directly to the objective function. The unconstrained, penalized surrogate objective is:

\begin{equation}\label{eq:trpo_penalty}
  \mathcal{J}^{\text{Penalty}}(\theta) = \hat{\mathbb{E}}_t \left[ \frac{\pi_\theta(a_t | s_t)}{\pi_{\theta_{\text{old}}}(a_t | s_t)} \hat{A}_t - \beta \, D_{\text{KL}}[\pi_{\theta}(\cdot | s_t) \| \pi_{\theta_{\text{old}}}(\cdot | s_t)] \right],
\end{equation}

where $\beta$ is a penalty coefficient that controls the trade-off between maximizing the expected advantage and minimizing the KL divergence (the negative sign before $\beta$). Selecting a suitable $\beta$ is difficult, and this motivates Proximal Policy Optimization (PPO), a first-order method that uses a clipped surrogate to approximate the trust region constraint efficiently \cite{schulman2017proximalpolicyoptimizationalgorithms}.

As shown in Eq.~\ref{eq:trpo_objective}, the KL-divergence constraint $\delta$ limits how much the new policy can deviate from the old policy in a single update, preventing destructive updates. In practice, the state distribution is unknown and approximated using trajectories collected under the old policy. It should be noted that TRPO uses constraints based on the mean KL divergence between successive policies. While TRPO successfully maintains this mean-KL trust region, this mean constraint does not successfully enforce the theoretically motivated maximum KL constraint~\cite{ilyas2018deep}. The maximum KL divergence is found to be several orders of magnitude larger than the enforced mean KL bound.

Directly optimizing $J(\theta)$ in Eq.~\ref{eq:vpg} via gradient-based methods is difficult because the environment dynamics $P(\tau \mid \pi_\theta)$ is  typically unknown and non-differentiable with respect to the policy parameters $\theta$. Moreoever, the trajectory distribution depends on $\theta$ through the stochastic sampling actions generated by the policy, making  $\theta J(\theta)$ only implicitly and non-differentiably dependent on $\theta$. As a result, the gradient $\nabla_\theta J(\theta)$ cannot be obtained by direct differentiation.

To address this, the policy gradient theorem provides a way to rewrite the gradient of the expected return in a computable form using the likelihood-ratio trick as in the original REINFORCE paper~\cite{williams1992simple}:

\begin{equation}
  \label{eq:vpg_gradient}
\nabla_\theta J(\theta) 
= \mathbb{E}_{\tau \sim \pi_\theta} \Big[ R(\tau) \, \nabla_\theta \log P(\tau \mid \pi_\theta) \Big]
= \mathbb{E}_{\tau \sim \pi_\theta} \Big[ \sum_{(s_t,a_t)\in \tau} R_t \, \nabla_\theta \log \pi_\theta(a_t \mid s_t) \Big],
\end{equation}

where $R_t = \sum_{t'=t}^{T} \gamma^{t'-t} r_{t'}$ denotes the discounted return accumulated from time step $t$.  Here we use the accumulated future reward $R_t$ after time step $t$ instead of the total return $R(\tau)$ for several important reasons. First, $R_t$ ensures that the policy gradient at step $t$ only takes into account rewards that come after the action $a_t$,  which avoids incorrectly assigning credit from rewards that happened before the action. Second, by focusing on rewards actually influenced by the action, we reduce noise in the gradient estimate. Finally, using $\gamma$ emphasizes near-term rewards more than distant ones. This makes learning more stable and aligns with how many real-world tasks care about immediate outcomes.

In Eq.~\ref{eq:vpg_gradient} the gradient depends only on the policy log-probabilities, which are differentiable with respect to $\theta$, allowing us to compute unbiased estimates of $\nabla_\theta J(\theta)$ from sampled trajectories. However, directly using $R_t$ as the weighting factor in the gradient estimator can result in high variance. To reduce variance, it is common to subtract a baseline $b(s_t)$ from the return~\cite{sutton1998reinforcement}, leading to the advantage function:

$$
A(s_t, a_t) = R_t - b(s_t),
$$

which preserves the expectation of the gradient while reducing its variance. In practice, the baseline is often chosen as the value function $V_{\pi_\theta}(s_t)=\mathbb{E}_{\pi_\theta}[R_t | s_t]$. With these modifications, the policy gradient can be expressed as:

\begin{equation}
\nabla_\theta J(\theta) 
= \mathbb{E}_{\tau \sim \pi_\theta} \left[ \sum_{(s_t,a_t)\in \tau}  \nabla_\theta \log \pi_\theta(a_t \mid s_t) \,  A(s_t, a_t) \right],
\end{equation}

Directly computing the expectation over all possible trajectories is generally intractable due to the combinatorial size of the state-action space. Consequently, policy gradient methods rely on empirical estimates computed from a finite set of sampled trajectories (rollouts) obtained by interacting with the environment. The most commonly used gradient estimator replaces the total return with the advantage function $\hat{A}_t$, yielding:

\begin{equation}
\hat{g} = \hat{\mathbb{E}}_t \left[ \nabla_\theta \log \pi_\theta(a_t \mid s_t) \, \hat{A}_t \right],
\end{equation}

where:  
\begin{itemize}
    \item $\hat{\mathbb{E}}_t[\cdot]$ denotes the empirical average over timesteps in the sampled dataset $\mathcal{D} = \{\tau_i\}_{i=1}^{N}$.
    \item $N$ is the total number of sampled trajectories.
    \item $\hat{A}_t$ is an estimator of the advantage function at time step $t$. In practice, the Generalized Advantage Estimator (GAE) \cite{schulman2015high} is often employed to reduce variance while maintaining low bias.
\end{itemize}

More explicitly, the policy gradient can be approximated by summing over all timesteps and trajectories in the dataset:

\begin{equation}
\hat{g} = \frac{1}{N} \sum_{\tau \in \mathcal{D}} \sum_{t=0}^{T} \nabla_\theta \log \pi_\theta(a_t \mid s_t) \, \hat{A}_t,
\end{equation}

This empirical estimate forms the basis for most modern policy gradient algorithms, including PPO and its variants.

\section{Advantage estimate and state-value function}

The advantage function measures how much better or worse an action performs relative to the critic’s current value estimate. In practice, it is computed using Generalized Advantage Estimation (GAE) to balance bias and variance in the policy gradient updates. GAE expresses the advantage as an exponentially weighted sum of temporal-difference (TD) errors:

\begin{equation}\label{eq:ppo_advantage}
\hat{A}_t = \sum_{l=0}^{\infty} (\gamma \lambda)^l  \delta_{t+l},
\end{equation}

where each TD error $\delta_t$ captures the discrepancy between the critic's current value prediction $V_t$ and the observed one-step return $r_t+\gamma V_{t+1}$:

\begin{equation}
\delta_t =\underbrace{r_t + \gamma V_{t+1}}_{\text{one-step return}} - \underbrace{V_t}_{\text{current value prediction}}
\end{equation}

Here, $\gamma$ is the discount factor, and $\lambda \in [0,1]$ is the GAE parameter controlling the trade-off between bias and variance.  Smaller values of $\lambda$ place greater weight on  short-term estimates from the critic, yielding low-variance but potentially higher bias, while larger $\lambda$ incorporates longer-term returns for reduced bias but increased variance. Setting $\lambda=1$ recovers the full Monte Carlo return, while $\lambda=0$ reduces to the one-step TD error.

The state-value function $V_t = V(s_t)$ is a core component of actor–critic methods such as PPO. It represents the expected future return—i.e., the cumulative discounted reward—that the agent can obtain from state $s_t$ when it continues to follow the current policy $\pi$. Formally, it is defined as the expected discounted sum of future rewards:

\begin{equation}\label{eq:ppo_critic_definition}
V_t = \mathbb{E}_{\pi}\left[\sum_{k=0}^\infty \gamma^k r_{t+k}\right].
\end{equation}

This expression can also be written recursively using the Bellman expectation equation, which relates the value of the current state to the immediate reward and the discounted value of the next state:

\begin{equation}\label{eq:ppo_critic_bellman}
V_t = \mathbb{E}_{\pi}[r_t + \gamma V_{t+1} \mid s_t].
\end{equation}

The recursive form highlights that the value function can be built up from an immediate reward plus the expected value of the next state, making it a fundamental building block for value-based learning and for constructing advantage estimates in PPO.

In PPO, the critic model $V_\phi(s_t)$ is trained to approximate the true value function $V_t$. Because the exact value $V_t$ is not directly observable, the critic relies on a bootstrapped target constructed from its own current value estimates:

$$
\hat{V}_t^{\text{target}} = \hat{A}_t + V_\phi(s_t).
$$

This target approximates the return while keeping the variance low. The critic is then optimizxed by minimizing a squared-error regression loss between its predictions and these targets:

\begin{equation} \label{eq:ppo_value_loss} 
L_V(\phi) = \mathbb{E}_t \Big[ \big( \hat{V}_t^{\text{target}} - V_\phi(s_t) \big)^2 \Big] 
\end{equation}

To further stabilize training, some PPO implementations apply clipping to the value prediction, analogous to policy clipping. The clipped value loss is:  
\begin{equation} \label{eq:ppo_value_loss_clipped} 
L_V^{\text{clip}}(\phi) =\mathbb{E}_t \Big[  
\max\Big(  
\big(V_\phi(s_t) - \hat{V}_t^{\text{target}}\big)^2,  
\big(\mathrm{clip}(V_\phi(s_t), V_{\phi_{\text{old}}}(s_t) - \epsilon, V_{\phi_{\text{old}}}(s_t) + \epsilon) - \hat{V}_t^{\text{target}}\big)^2  
\Big)  
\Big].  
\end{equation}

These learned value predictions are then used in the computation of the advantage $\hat{A}_t$.

To better understand the advantage function,  we can expand Eq.~\ref{eq:ppo_advantage} by substituting the definition of $\delta_t$. This gives:

 $$
\begin{aligned}  
\hat{A}_t  
&= [r_t + \gamma\lambda r_{t+1} + (\gamma\lambda)^2 r_{t+2} + \dots] \\  
&\quad + [\gamma V_{t+1} - V_t
- \gamma\lambda(\gamma V_{t+2} - V_{t+1})
- (\gamma\lambda)^2(\gamma V_{t+3} - V_{t+2}) + \dots] 
\end{aligned}
$$

The first bracket is a discounted sum of future rewards, $R_t$. The second bracket is a telescoping series that simplifies to:

\begin{equation}\label{eq:telescoping}
-V_t + (1 - \lambda)\sum_{l=1}^{\infty} (\gamma \lambda)^{l-1} \gamma^l V_{t+l}   
\end{equation}

This lets us rewrite the advantage function as:
\begin{equation}\label{eq:advantage_decomposition}
\hat{A}_t = R_t -V_t + (1 - \lambda)\sum_{l=1}^{\infty} (\gamma \lambda)^{l-1} \gamma^l V_{t+l}   
\end{equation}

When $\lambda \to 1$, the additional sum term vanishes, giving:

$$
\hat{A}_t \approx R_t  -V_t  
$$


\end{document}